%% file: main.tex
\documentclass{article} 
\usepackage{authblk}
\usepackage{macros}
\usepackage{arxiv}

\newtoggle{arxiv}
\toggletrue{arxiv}

\usepackage{csquotes} 
\usepackage{url}            
\usepackage{graphicx}
\usepackage[numbers,sort]{natbib}
\usepackage{subfiles}
\usepackage{tabularx}
\usepackage{etoolbox}
\usepackage{multirow}
\usepackage{enumitem}
\usepackage{amssymb}
\usepackage{amsmath}
\usepackage{amsthm}
\usepackage{xspace}
\usepackage{float}
\usepackage{wrapfig}
\usepackage{booktabs}       
\usepackage{amsfonts}       
\usepackage{nicefrac}       
\usepackage{microtype}      
\usepackage{color, soul}
\usepackage{pifont}
\usepackage{caption}
\usepackage{subcaption}
\usepackage{nth}
\usepackage{listings}
\usepackage{xcolor}
\usepackage{seqsplit} 
 
\lstdefinestyle{livepatch}{
  language=Python,
  basicstyle=\ttfamily\footnotesize,
  commentstyle=\color{gray!70!black}\itshape,
  keywordstyle=\color{blue!60!black},
  stringstyle=\color{teal!70!black},
  showstringspaces=false,
  columns=fullflexible,
  keepspaces=true,
  breaklines=false,
  xleftmargin=1em,
  frame=single,
  framerule=0.4pt,
  rulecolor=\color{gray!50},
  morekeywords={def,return,if,not},
}

\usepackage[framemethod=TikZ]{mdframed}

\usepackage{listings}
\definecolor{codegreen}{rgb}{0,0.6,0}
\definecolor{codegray}{rgb}{0.5,0.5,0.5}
\definecolor{codepurple}{rgb}{0.58,0,0.82}
\definecolor{backcolour}{rgb}{1,1,1}

\usepackage{listings}  
\usepackage{algorithm}
\algrenewcommand\algorithmicrequire{\textbf{Input:}}
\algrenewcommand\algorithmicensure{\textbf{Output:}}

\usepackage[table]{xcolor}

\newcommand{\blfootnote}[1]{%
  \begingroup
  \renewcommand{\thefootnote}{}%
  \footnotetext{#1}%
  \addtocounter{footnote}{-1}%
  \endgroup
}

\iftoggle{arxiv}{
    \setlength{\textwidth}{6.5in}
    \setlength{\textheight}{9in}
    \setlength{\oddsidemargin}{0in}
    \setlength{\evensidemargin}{0in}
    \setlength{\topmargin}{-0.5in}
    \newlength{\defbaselineskip}
    \setlength{\defbaselineskip}{\baselineskip}
    \setlength{\marginparwidth}{0.8in}
}

\newcolumntype{Y}{>{\hsize=.7\hsize}X}
\newcolumntype{Z}{>{\hsize=1.3\hsize}X}

\makeatletter
\def\@copyrightspace{\relax}
\makeatother

\makeatletter
\def\@myauthornotes{}
\def\myauthornote#1{%
  \if@ACM@anonymous\else
    \g@addto@macro\addresses{}%
    \g@addto@macro\@myauthornotes{%
      \stepcounter{footnote}\footnotetext{#1}}%
  \fi}
\makeatother

\iftoggle{arxiv}{
    \renewcommand{\thefootnote}{}
    \title{Together Compiler: Continual Compilation for AI Inference}
    \usepackage{authblk}
    \setlength{\affilsep}{10pt}
    \author{\vspace{10pt}
            Reyna Abhyankar\textsuperscript{*}
            \hspace{0.5em} Dylan Lim\textsuperscript{*}
            \\[3pt]
            Drew Wadsworth \hspace{0.5em}
            Venkata Hasith Vattikuti \hspace{0.5em}
            YuTong Julia Zhang \hspace{0.5em}
            Jung Hwan Heo
            \\[3pt]
            Taeksang Kim \hspace{0.5em}
            Kesavan Ramakrishnan \hspace{0.5em}
            Evangelene Stanley \hspace{0.5em}
            Xinyu Fang \hspace{0.5em}
            Daniel Y. Fu
            \\[3pt]
            Simran Arora
    }
    \affil{Together AI Research}
    \vspace{4pt}
}

\definecolor{dkgreen}{rgb}{0,0.6,0}
\definecolor{gray}{rgb}{0.5,0.5,0.5}
\definecolor{light-gray}{gray}{0.95}
\definecolor{mauve}{rgb}{0.58,0,0.82}
\definecolor{backcolour}{rgb}{0.95,0.95,0.92}

\usepackage{graphicx}
\usepackage{subcaption}
\usepackage{wrapfig,lipsum,booktabs}
\usepackage{threeparttable}
\usepackage{makecell}
\usepackage{hyperref}
\usepackage{url}
\usepackage{algorithm}
\usepackage{algpseudocode}
\usepackage{hyperref}
\usepackage{cleveref}
\usepackage{changepage}
\usepackage{booktabs}

\usepackage{epstopdf}
\usepackage{graphicx}
\usepackage{textcomp}       
\usepackage{scalerel}       
\newcommand{\name}{\textsc{Together Compiler}}
\newcommand{\shortname}{\textsc{TKCC}}

\usepackage{listings}
\usepackage{color}
\definecolor{dkgreen}{rgb}{0,0.6,0}
\definecolor{gray}{rgb}{0.5,0.5,0.5}
\definecolor{mauve}{rgb}{0.58,0,0.82}
\lstdefinelanguage{CUDACPP}{
  language=C++,
  morekeywords={__global__, __host__, __device__, __shared__, blockIdx, blockDim, threadIdx, gridDim},
  morecomment=[l][\color{magenta}]{\#},
}
\lstdefinestyle{pythonstyle}{
  language=Python,
  frame=tb,
  aboveskip=3mm,
  belowskip=3mm,
  showstringspaces=false,
  columns=flexible,
  basicstyle={\ttfamily\footnotesize},
  numbers=none,
  numberstyle=\footnotesize\color{gray},
  keywordstyle=\color[rgb]{0.13,0.29,0.53},
  commentstyle=\color[rgb]{0.13,0.55,0.13},
  stringstyle=\color[rgb]{0.31,0.60,0.02},
  breaklines=true,
  breakatwhitespace=true,
  tabsize=4
}

\input{math_commands.tex}

\usepackage{enumitem}

\newcommand{\ShowNotes}{}
\input{macros}

\begin{document}

\maketitle

\blfootnote{%
  \mbox{%
    \textsuperscript{*}Equal contribution.
  }%
}
\blfootnote{Correspondence to {\hypersetup{urlcolor=black}\href{mailto:rabhyankar@together.ai,dylanlim@together.ai,simran@together.ai}{\texttt{\{rabhyankar,\,dylanlim,\,simran\}@together.ai}}}.}

\input{Sections/0_abstract}

\section{Introduction}
\label{Introduction}
\input{Sections/1_intro}

\section{Background}
\input{Sections/2_background}

\section{Motivation for Continual Compilation}
\input{Sections/3_motivation}

\section{Method}
\input{Sections/4_method}

\section{Results}
\input{Sections/5_results}

\section{Conclusion}
\input{Sections/6_conclusion}

\section{Acknowledgments}
We thank Ce Zhang, Chris Ré, Neel Guha, Stuart Sul, Ali Yassine, Alex Angus, Austin Silveria, Gordon Gustafson, Kevin Cui, Michael Granado, Sonny Li, Xinyi Li, Peter Wu, Sanjana Srivastava, Federico Bianchi, Pablo Cases, Susan Lee, and Nikitha Suryadevara for helpful feedback and discussions during this work.

\bibliography{main}
\bibliographystyle{main}
\appendix
\clearpage

\input{Sections/appendix/7_appendix}

\vspace{1mm}

\end{document}

%% file: math_commands.tex
\usepackage{amsmath,amsfonts,bm}

\def\eqref#1{equation~\ref{#1}}

\def\1{\bm{1}}

\DeclareMathAlphabet{\mathsfit}{\encodingdefault}{\sfdefault}{m}{sl}
\SetMathAlphabet{\mathsfit}{bold}{\encodingdefault}{\sfdefault}{bx}{n}



%% file: macros.tex
\ifdefined\ShowNotes
  \newcommand{\colornote}[3]{{\color{#1}\bf{#2 #3}\normalfont}}
\else
  \newcommand{\colornote}[3]{}
\fi

\definecolor{darkred}{rgb}{0.7,0.1,0.1}
\definecolor{darkgreen}{rgb}{0.1,0.5,0.1}
\definecolor{cyan}{rgb}{0.7,0.0,0.7}
\definecolor{dblue}{rgb}{0.2,0.2,0.8}
\definecolor{maroon}{rgb}{0.76,.13,.28}
\definecolor{burntorange}{rgb}{0.81,.33,0}
\definecolor{royalpurple}{rgb}{0.47,.31,0.66}

\ifdefined\ShowNotes
  
\else
  
\fi

%% file: Sections/0_abstract.tex
Today's AI workloads are rapidly changing: new models are released more often, model architectures are increasingly diverse, and traffic distributions change frequently. The challenge we face at Together AI is continual optimization for these changing workloads. Compiling a given deployment into an optimized forward pass is a challenge that is well-suited for agent hill-climbing. Existing systems struggle with continual compilation due to a tradeoff between measurement throughput and fidelity. Many systems use agents to optimize individual kernels from the inference forward pass using offline benchmarks, but timings are low-fidelity because of data-dependent behavior in primitives like mixture-of-experts (MoE). On the other hand, systems use agents to directly optimize end-to-end engines but suffer from slow engine launch times. To address these challenges, we build \name{} (\shortname{}), the first \textit{truly continual} agent loop designed for AI inference optimization. To address the throughput-fidelity tradeoff, we propose: (1) a Tiered Persistent Engine that decouples engine launch overheads from forward pass computation; (2) a cost model to select the parallelism configurations that seed the optimization process. We also conduct extensive ablations on agent context management and present our insights on deploying \shortname{} in production. \shortname{} outperforms highly-optimized hand-written Megakernels on the Llama 1B workhorse workload by up to $1.7\times$ and the best open source engines on the latest frontier models (Kimi K3, GLM 5.3, MiniMax M3.1 Flash, DeepSeek v4.1 Flash) across hardware platforms by up to $2.6\times$ higher tokens per second (TPS) on SemiAnalysis' InferenceX benchmark. \shortname{} also sets the state-of-the-art performance on low-latency, batch size 1 inference, generating 1111 TPS on MiniMax M3.1 Flash and 1205 TPS on DeepSeek v4.1 Flash. Finally, \shortname{} generalizes and provides speedup beyond frontier language model inference. \shortname{} serves trillions of tokens across our production fleet.

%% file: Sections/1_intro.tex
When AI agents have been presented with well-specified targets in mathematics, engineering, and science~\cite{openai2026math, openai2026navierstokes}, they have been effective at hill-climbing to achieve their objective.
The key challenge for inference optimization at Together AI is that our traffic shape, i.e. our target to optimize, is constantly evolving. We explore the problem of \textit{continually} compiling and optimizing our inference stack using AI agents.

Traffic primarily evolves due to three reasons. The first reason is behavioral trends. For instance, Lab A's model is more heavily utilized for agent-shaped workloads in one country compared to the rest of the world, which uses it for chat-shaped workloads. The second reason is burstiness, where a customer suddenly increases its usage or changes its use case. The third reason is shifts in usage patterns as users learn more about a model's capabilities. For Lab
B’s model, traffic gradually moved towards longer context, from 10K to 18K, and cache hit jumped by 8 percentage points in the 2-week period following the model's release.

There are many existing recent agentic optimization systems, but they either tackle narrow tasks (e.g., hyperparameter search~\cite{chen2026autoscalingautoresearch} or one-off kernel benchmarking~\cite{feng2026krexconcurrentkernelbenchmarking,lin2026multiagentkernels}) or their design decisions have not been thoroughly examined across production traffic patterns~\cite{liu2026rsiinference,glm5team2026glm5,wang2025geak}. When stress-testing existing systems on production traffic, we find that they are unable to achieve strong performance because they face a tradeoff between insufficient optimization throughput and insufficient measurement fidelity: 
\begin{itemize}
    \item \textbf{Low fidelity, high throughput.} Many agent systems optimize individual AI kernels on standalone benchmarks using random input data~\cite{ouyang2025kernelbench, lin2026solexecbench}. However, this hurts measurement fidelity since the execution behavior of modern AI architectures is sensitive to the input data distribution (e.g., mixture-of-experts,  sparse top-k attention)~\cite{kimik32026, zai2026glm53, deepseekai2026deepseekv41flash}. 
    In \Cref{sec:motivation}, we show that measurement fidelity can impact the ideal optimization strategy for mixture-of-experts (MoE) kernels.
    \item \textbf{High fidelity, low throughput.} Other recent systems apply agents to end-to-end inference engine optimization, where the engine loads real model weights and measures speedups on real input data~\cite{liu2026rsiinference, wang2025geak}. This uses a high fidelity data distribution but suffers from painfully slow experiment iteration times due to engine launch overheads that take minutes (\Cref{method:engine}).
    All this idle GPU time is especially wasteful given the rising cost of AI compute~\cite{flight2026tokenomics}.
\end{itemize}

\input{Figures/1_1_overview}

Ideally, we could enable high-throughput iteration while preserving measurement fidelity. We introduce {\name{}} ({\shortname{}}), the first \textit{truly continual} agent-driven loop for AI inference optimization that produces fast forward passes and integrates them directly into the serving engine (Figure~\ref{fig:arch-overview}). \shortname{} uses agents to search for optimizations and balances the throughput-fidelity tradeoff to unlock speedups. 
We highlight three key contributions from deploying {\shortname{}} in production:

\begin{enumerate}
\item \textbf{Tiered Persistent Engine.} Agents need high-fidelity measurements, but obtaining them requires expensive inference engine start up. To address this tradeoff between measurement throughput and fidelity, we introduce a new paradigm for iterating on inference optimization: the Tiered Persistent Engine. Our finding is that engine startup can be decoupled from forward pass computation, and the forward pass can be \textit{hot-reloaded} with minimal overhead. {\shortname{}} initializes the production engine once and establishes two measurement tiers: a fast tier for high-throughput iteration and a slow tier for high-fidelity signal. The Tiered Persistent Engine allows \shortname{} to process 252$\times$ more agent experiments than existing inference optimization systems (\Cref{method:engine}).

\item \textbf{Cost model for parallelism selection.} As inference architectures and traffic evolve, selecting the optimal parallelism configuration requires balancing prediction fidelity against search cost. Two sources of uncertainty complicate this process: data-dependent execution (e.g., sparse KV accesses, MoE expert imbalance, and variable request lengths) and the maturity of existing software implementations, which can obscure hardware performance potential. We propose a cost model that ranks the top-$k$ parallelism configurations using analytical estimates and tiered measurements, ranging from theoretical hardware capabilities to measured compute, memory, and communication performance. This enables \shortname{} to identify promising configurations before investing in expensive agentic optimization (\Cref{method:search}).

\item \textbf{Bitter lessons~\cite{sutton2019bitterlesson} of performance optimization.} 
We have been studying frameworks for accelerating AI inference for the past 2 years, starting from creating ThunderKittens~\cite{spector2024tk} and Megakernels~\cite{spector2025look, spector2025we} to building an in-house compiler. Much of our investigation predates tools like Claude Code, \texttt{/goal} or \texttt{auto mode}, or high-quality compaction. Earlier versions of \shortname{} deployed in production utilized rule-based lowering, built custom intermediate representations, and explicitly managed agent context. We continued to ablate these decisions as coding harnesses improved and have observed that agents without abstractions can achieve a higher performance ceiling (\Cref{method:lessons}).

\end{enumerate}

We evaluate \name{} on a variety of workloads against state-of-the-art baselines and make three primary claims, detailed in \Cref{sec:sec5_results}. First, \name{} outperforms highly-optimized megakernel baselines (e.g., ThunderKittens~\cite{spector2024tk} and Mirage Persistent Kernel~\cite{cheng2026mpk}) on their workhorse workload: up to $1.7\times$ higher tokens per second (TPS) on Llama 1B (2,100 TPS). Second, \name{} adapts to different models and engine configurations, outperforming the best open performance on the SemiAnalysis InferenceX benchmark~\cite{semianalysis2026inferencex} across frontier models (Moonshot Kimi K3~\cite{kimik32026}, Z.ai GLM 5.3~\cite{zai2026glm53}, MiniMax M3.1 Flash~\cite{minimaxm32036}, DeepSeek v4.1 Flash~\cite{deepseekai2026deepseekv41flash}) for diverse hardware platforms by up to $2.6\times$ higher TPS. This was achieved in under 24 hours using at most 2 nodes for each model. We also set the state-of-the-art performance on low-latency batch size 1 inference, achieving 1,111 TPS on B200 M3.1 Flash and 1,205 TPS on GB300 DeepSeek v4.1 Flash. Third, {\shortname{}} generalizes to speedup in domains beyond language modeling. \name{} serves trillions of tokens per day and continually compiles our production fleet.

%% file: Figures/1_1_overview.tex
\begin{figure*}[!t]
    \centering
     \includegraphics[width=\textwidth]{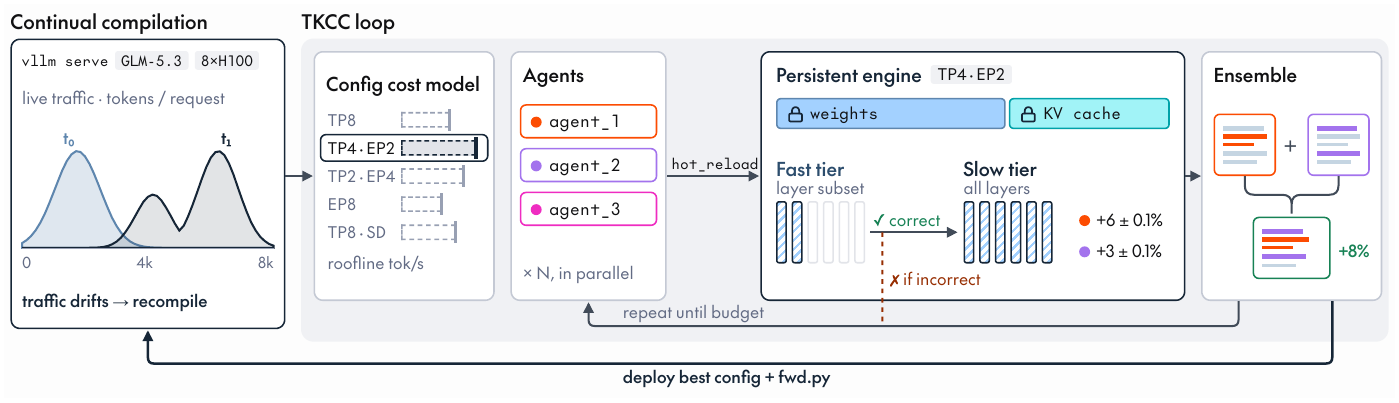}
    \caption{\textbf{Together Compiler architecture.} As input, \shortname{} takes in a server launch command, traffic profile, and hardware configuration. It uses a cost model to select the starting parallelism configuration for optimization, spawns parallel agents that are constantly communicating, and measures generated optimizations with a persistent engine. The final source code is directly integrated into the serving engine. As the traffic profile shifts over time, \shortname{} continually adapts.}
    \label{fig:arch-overview}
\end{figure*}

%% file: Sections/2_background.tex
\label{sec2_preliminaries}
We provide background on the following work: emerging model architectures, inference engine deployments, agentic kernel generation, and agent-driven optimization loops.

\paragraph{Emerging model architectures.} Recent frontier open models -- Z.ai's GLM 5.3~\cite{zai2026glm53, glm5team2026glm5}, DeepSeek v4.1 Flash~\cite{deepseekai2026deepseekv41flash}, MiniMax M3.1 Flash~\cite{minimaxm32036} -- have started using two architectural patterns: MoE and sparse attention. MoE models dynamically route each query token to a subset of the model weights~\cite{shazeer2017outrageously, fedus2022switch}, as opposed to dense transformers that apply the full model weights to every token~\cite{vaswani2018attention}. MoE models thus maintains a large parameter space during training with higher inference efficiency~\cite{fedus2022switch, dai2024deepseekmoe, deepseekai2024deepseekv3}. Sparse attention dynamically selects and attends to a subset of the sequence~\cite{child2019generating, yuan2025nsa, lu2025moba, deepseekai2025deepseekv32}, as opposed to non-sparse attention that attends to all tokens~\cite{vaswani2018attention}. Sparse attention improves inference efficiency at higher context lengths~\cite{yuan2025nsa, deepseekai2025deepseekv32}. 

\paragraph{Inference deployments.} An inference engine schedules requests to be executed by the model's forward pass code. Requests can differ in their input sequence length (ISL) and output sequence length (OSL). Inference engines can choose to batch requests together~\cite{kwon2023vllm} for higher throughput. Frontier models are served across several GPUs with different modes of parallelism~\cite{shoeybi2019megatron}, which result in a different layout for weights and KV cache data. TensorRT-LLM~\cite{trt}, SGLang~\cite{zheng2024sglang}, and vLLM~\cite{kwon2023vllm} are examples of engines that differ in batching, cache layout, and model implementations.

\paragraph{Agentic kernel generation.} Coding agents are rapidly improving at kernel generation~\cite{ouyang2025kernelbench, yu2026kernelgensurvey} and now beat expert baselines~\cite{luo2026agentickernelopt}. Agents can author kernels in raw CUDA/PTX, a domain-specific language (DSL), or even a custom intermediate representation (IR) designed specifically for agents, such as TIRx and CAKE~\cite{tvm2026tirx, ye2026cakecompiler}. Systems built around agents (GEAK, Astra, STARK, KernelAgent, AutoComp, AccelOpt)~\cite{wang2025geak, wei2025astra, dong2025stark, pytorch2026kernelagent, hong2025autocomppowerfulportablecode, zhang2026acceloptselfimprovingllmagentic} typically employ the following strategy for choosing which kernels to optimize: they rank kernels with a cost model based on roofline or wall-clock estimates and then iteratively measure and refine one kernel at a time. We find that explicit strategies can negatively bias agent search (\Cref{method:lessons}), while individual kernel optimization may not translate to end-to-end speedup (\Cref{sec:motivation}).

\paragraph{Agent-driven loops for inference optimization.} 

Optimization loops involve agents proposing solutions, collecting measurements, and integrating feedback~\cite{openai2026math, openai2026navierstokes}. Recent systems have built such loops for inference optimization~\cite{zai2026glm53, decagon2026autoresearch}. Apple's RSI-Inference Bench~\cite{liu2026rsiinference} evaluates different coding agents on a variety of inference optimization tasks in vLLM and SGLang. We find that these systems face low agent experiment throughput and GPU utilization because they launch the entire inference engine for each measurement of a generated optimization. We explore this in the following section (\Cref{sec:motivation}).

%% file: Sections/3_motivation.tex
\label{sec:motivation}

Together AI serves many trillions of tokens per day across thousands of customers, so our services are constantly experiencing variable traffic shapes. The following are the primary dimensions of variation: 

\begin{itemize}
    \item \textbf{Behavioral trends.} Traffic shape is impacted by repeated patterns. Some countries use Lab A's model for agentic traffic, while other countries use the same model for chat. This results in the ISL shifting from $\approx 97$k during one country's working hours to an ISL of $\approx 70$k during off-hours.
    \item \textbf{Burstiness.} Customers frequently shift their requests per second or use case, impacting cache hit rate, ISL, and OSL. For example, within a period of 3 days, Customer A's P90 ISL grew from 18k to 33k tokens, OSL grew from 1k to 1.9k, and cache hit rate fell 4 percentage points. For Customer B, the average OSL increased from 700 to 1.7k, and cache hit rate fell by 7 percentage points within the span of one week.
    \item \textbf{Gradual shift.}  Model labs roll out new checkpoints to our team over the course of their training runs, which can induce changes in the traffic shape. For example, Lab A rolled out a new checkpoint that generated more tokens during its thinking phase, resulting in $2\times$ longer outputs. For Lab B's model, traffic gradually moved towards longer context, from 10K to 18K, and cache hit rate jumped by 8\% over a 2-week period as customers shifted from evaluating the model to adopting the model.
\end{itemize}

Such changes in traffic requires continual optimization to maximize the GPU utilization of our fleet. 

\paragraph{Throughput and fidelity tradeoffs.} Recently, coding agents have provided significant value in accelerating performance optimization. Maximizing speedup requires high experimentation throughput and high-fidelity measurements: 

\begin{itemize}
\item \textbf{Low fidelity, high throughput.} Existing agent systems from companies like Alibaba~\cite{feng2026krexconcurrentkernelbenchmarking} and Cursor~\cite{lin2026multiagentkernels} iterate on standalone kernels in offline benchmark harnesses outside the inference engine (e.g., KernelBench~\cite{ouyang2025kernelbench}), often using random data as input. It is fast for agents to quickly try many optimizations using these harnesses, but random inputs provide low-fidelity measurements.

    \begin{itemize}
    \item Randomly initialized data activates 13-38\% more experts than real traffic in \Cref{fig:fid-experts}. On DeepSeek v4.1 Flash, we observe that Cursor's production-grade MoE megakernel system (Mixture-of-Kittens)~\cite{sul2026mok} incurs a $6$-$15\%$ slowdown when measured on real traffic compared to balanced load across experts~\cite{sul2026mok} (\Cref{fig:fidelity}). 
    \item We also observe a similar problem with the indexer operation in sparse attention. An agent-authored top-$k$ kernel was written assuming random data for its inputs, but this approach scaled quadratically on real data, which caused a 1.3$\times$ slowdown end-to-end in the engine (\Cref{fig:fid-topk}). 
    \end{itemize}
    
    \item \textbf{Higher fidelity, low throughput.} Other AI systems from companies like AMD~\cite{wang2025geak}, Apple~\cite{liu2026rsiinference}, and Decagon~\cite{decagon2026autoresearch}, focus on end-to-end inference optimization. However, we find that they are unable to keep up with the velocity of traffic shifts. This is because their agents' experiments are bottlenecked by over ten minutes of inference engine startup time (\Cref{tab:m31flash}). 
\end{itemize}

Next, we introduce {\name{}}, the first \textit{truly continual} agentic loop for AI inference optimization, powered by high-throughput and high-fidelity design decisions to maximize speedup.

%% file: Sections/4_method.tex
In this section, we describe {\name{}}'s design and our lessons from developing the system. First, we propose a novel Tiered Persistent Engine architecture that provides high-fidelity and high-throughput measurement feedback to agents. Second, we propose a cost model that ranks the top-$k$ parallelism configurations and selects the starting point for optimization. Third, we share insights from the previous two years of studying frameworks for accelerating AI inference and find that the bitter lesson~\cite{sutton2019bitterlesson} holds true for performance optimization.

\subsection{System overview}
Figure~\ref{fig:arch-overview} shows the system architecture. \shortname{} takes as input the traffic shape, model, and hardware. \shortname{}'s goal is to use agents to generate a series of optimizations for the inference engine and validate them by collecting measurements. \shortname{} spins up a persistent engine that loads the model once and accepts any agent-generated forward pass as a ``candidate.'' \shortname{} uses a custom cost model to select the parallelism strategy for the persistent engine. The persistent engine \textit{hot-reloads} a candidate and runs it through different measurement tiers that are tailored towards a traffic profile. To maximize GPU utilization, \shortname{} uses multiple parallel agents that are constantly communicating with each other to generate many diverse candidates. \shortname{} perpetually loops until a given budget is exhausted, at which point we run a full suite of verification tests (e.g., for KV cache contamination and quality), ensuring that generated candidate optimizations can be served directly in production.

\subsection{High-throughput Tiered Persistent Engine} 
\label{method:engine}

To apply \shortname{} at production scale, minimizing GPU idle time is critical. We propose to achieve high GPU utilization during optimization by constructing a persistent engine with distinct measurement tiers. This design leads to high agent experiment throughput and measurement fidelity.

\input{Tables/tab_4_3_engine_startup_arxiv}

\input{Figures/3_1_tiered_speedup}
\paragraph{Persistent engine with hot reloading.} We observe that engine launch and teardown can be decoupled from model forward pass execution. Based on this observation, we build a persistent engine that can run different forward pass variants without restarting the engine. This is a new paradigm for agentic inference optimization; \shortname{} enables 252$\times$ higher throughput for agents and 11$\times$ lower idling time per agent compared to a system that restarts the engine for each measurement.

\input{Sections/_patch_engine_code_snippet}
Figure~\ref{fig:live-patch} shows our implementation of ``hot reloading." The engine is initialized once with an additional endpoint that accepts a candidate forward pass. When an agent submits its candidate as a ``job" to the endpoint, the server replaces the model’s forward function pointer and re-captures CUDA/HIP graphs. To ensure GPUs are fully utilized, \shortname{} spawns parallel agents so that one agent can think and make tool calls while other agents are receiving measurement feedback. 

Two further optimizations to improve throughput are (1) Prefix cache reuse: rather than using fresh requests for each measurement, \shortname{} reuses a fixed set of requests. Since the prompts remain cached across measurements, measurement latency reduces by an additional $2.6\times$ on Kimi K3. 
(2) Binary reuse: many kernels remain the same from one submission to the next, so we cache kernel binaries to prevent needless recompilation. 

Given the high-throughput measurement framework, agents must use the experimentation capacity effectively. We find that maximizing agent communication is critical for avoiding redundant tasks, improving optimization speed by $2.3\times$ (Figure~\ref{fig:search_strats}).

\paragraph{Tiered measurement.} \shortname{} instantiates multiple persistent engines as distinct measurement tiers: a high-throughput tier and a high-fidelity tier. Agents submit to the faster tier to weed out slower kernels that could cause illegal memory accesses (IMAs) or hangs to avoid expensive engine restarts (Table~\ref{tab:m31flash-boot}). Slower tiers contain more layers (Table~\ref{tab:m31flash-tiers}) and provide higher-fidelity measurements for data-dependent architectures.
Agents see each tier's queue depth and job completion times and can load-balance among themselves. The fast tier also traces kernels (e.g., with CUPTI on NVIDIA) to provide agents with kernel-level reports.
In Figure~\ref{fig:tiered_speedup}, we show the two-tiered measurement system evaluates 8.1$\times$ more candidates than a single slow tier (1{,}354 vs.\ 168) and finds a solution that is 7.2\% faster. 

We find that running the same forward pass back-to-back in the engine can vary measurements by 4-6\%, which is problematic because a valid optimization can boost performance by even as little as $1\%$.
To reduce variance, we (1) Split prefill and decode: with continuous batching, decode timings vary based on whether it is mixed with prefill requests. We decouple prefill from decode measurements, reducing variance in back-to-back runs from 6.6\% to 2.7\%. (2) Tune measured layers: We find that the number of measured layers affects measurement variance. Before optimization starts, we tune the minimal number of layers needed for representative measurements for that particular workload.

\subsection{Cost model for configuration search}
\label{method:search}

\input{Figures/cost_model_crossover/cost_model_crossover}

We find that changes in traffic shape may also need changes to the inference engine's parallelism configuration. Therefore, we introduce a cost model to identify the top-$k$ promising configurations before investing in expensive agent optimization.

Existing systems have developed sophisticated methods for selecting inference configurations under SLA requirements~\cite{zhong2024distserve, xu2026aiconfigurator, pope2023efficiently, zhu2025megascaleinfer}. For continual configuration selection, we face a tradeoff between prediction fidelity and search cost: measuring every configuration is expensive and reflects the immaturity of existing software implementations, while simple analytical models can miss data-dependent execution effects such as MoE imbalance, multi-token prediction (MTP) acceptance lengths, and uneven request lengths in a batch. We address these challenges with a cost model that combines analytical estimates with tiered hardware performance measurements.

\paragraph{Cost modeling.} We construct the search space for possible configurations, prune the search space, and collect high-fidelity measurements. Then, we use our model to provide a throughput estimate for the configuration at a fixed interactivity SLA.
\begin{enumerate}
    \item \textbf{Search space.} The search space considers replica sizes from one to a fixed maximum number of chips (e.g., 4096).  We collect all combinations across: \textbf{MoE}: Expert Parallel (EP), Tensor Parallel (TP); \textbf{Dense}: Tensor Parallel; \textbf{Attention}: Tensor Parallel, Data Parallel (DP), Context Parallel (CP).
    \item \textbf{Pruning.} We prune configurations based on a few factors. First, we eliminate configurations that cannot fit the model weights or evenly divide any of the model's tensor dimensions (e.g., the number of attention heads should be divisible by the number of chips). Second, we eliminate configurations if they cannot meet a fixed prefill time-to-first-token (TTFT) and decode TPS SLAs (1.5s TTFT, 40 TPS). For example, layouts that are spread over hundreds of nodes are often eliminated due to high communication overhead. The remaining rules are described in Appendix~\ref{app:cost-model}. 
    \item \textbf{Tiered measurements.} We account for two sources of uncertainty when estimating configuration performance: the gap between theoretical hardware capabilities and current software performance, and variability in production traffic.
    \begin{enumerate}
        \item First, configurations should not be penalized solely because their current implementations are poorly optimized. We therefore distinguish between three levels of performance estimates: theoretical hardware specifications, measured compute/memory/collective performance, and measured end-to-end forward-pass performance. Comparing these estimates helps distinguish hardware limitations from software implementation overheads (Figure~\ref{fig:k3-specs}). 
        \item Second, several components of modern inference are data-distribution dependent (e.g., MTP acceptance rates~\cite{liu2026speculative} and MoE expert activation imbalance~\cite{muennighoff2024olmoe}) and workload dependent (e.g., prompt unevenness across requests in a batch). We measure these distributions in production, since uneven expert routing and request lengths can create stragglers that determine the execution time of an entire parallel replica. 
    \end{enumerate} 
    \item \textbf{Model.} Given a model, a chip, and a set of parallelism configuration candidates, the cost model estimates the forward-pass time ($T_{\text{pass}}$) of the prefill and decode phases per unique model layer. Each layer's time is the larger of its compute time ($T_c^L$, from FLOPs and peak throughput) and its memory time ($D^L/BW$, from bytes read and memory bandwidth), plus its collective time ($T_{\text{comm}}^L$, from bytes moved and interconnect bandwidth and latency). The model then finds the largest batch ($C$) that fits in high-bandwidth memory (HBM) alongside weights and KV cache ($C_{\text{memory}}$) while still meeting the phase's latency target ($C_{\text{speed}}$), giving chip-seconds per request for that phase ($\text{chip-s}_{\text{prefill}}$, $\text{chip-s}_{\text{decode}}$). Prefill and decode chip-seconds are summed and inverted into a single metric: tokens per minute per chip ($\text{TPM}$).
  \begin{equation*}
  \begin{aligned}
    T_{\text{pass}} &= \sum_{L} \Big[\max\!\Big(T_c^L,\ \tfrac{D^L}{BW}\Big) + T_{\text{comm}}^L\Big] \\[4pt]
    \tau_{\text{pass}} &= \dfrac{N\, T_{\text{pass}}\, n_{\text{pass}}}{C} \\[4pt]
    \text{TPM} &= \dfrac{60\,(\text{ISL}+\text{OSL})}{\tau_P + \tau_D}
  \end{aligned}
  \end{equation*}
\end{enumerate}

\paragraph{Using the cost model.}
We illustrate how the cost model selects different configuration as model architectures, latency requirements, and usage patterns evolve. \textit{Architecture:} For models with fewer experts, such as GLM 5.3, smaller TP configurations are favored because additional parallelism introduces communication overhead without sufficient computational benefit. As expert counts increase, as in Kimi K3, wide EP configurations spanning 32 or 64 GPUs become more efficient. \textit{Latency requirements:} For Kimi K3, wide EP is favored in throughput-oriented regimes, where higher concurrency amortizes communication costs. As the per-user TPS requirement increases, the maximum feasible concurrency decreases, favoring higher TP degrees for both attention and MoE (Figure~\ref{fig:k3-decode}). \textit{Usage patterns:} As the MTP acceptance length increases, each decoding step generates more tokens, relaxing the latency constraint for a fixed TPS target. This allows for more replicas to support higher concurrencies, shifting the optimal configuration toward 64-GPU wide EP rather than 32-GPU configurations.

The cost model was able to improve our production deployment for Kimi K3 by finding a configuration that boosted TPS by up to 24\% and TPM by up to 18\%. In \Cref{fig:tkcc-c1-results-p90}, we use this cost model to select parallelism configurations that lead to speedups over InferenceX parallelism configs to improve per-user interactivity metrics.

\input{Figures/3_3_search}

\subsection{Lessons from building the {\name{}}}
\label{method:lessons}

Our primary takeaway is that broad AI exploration can now increasingly outperform prescriptive intermediate representations, curated optimization strategies, and explicit agent context management. 

\paragraph{System versions.} We have studied frameworks for AI inference acceleration for the past two years, putting three distinct iterations of the design into production for customer use:
\begin{enumerate}
    \item \textbf{Megakernels~\cite{spector2025look, spector2025we}.} We fuse all model operations into a single kernel to minimize launch overheads and tail effects. Specialized warps perform instruction scheduling on the device side.
    \item \textbf{Torch Compile~\cite{ansel2024pytorch2} augmentation.} Given the difficulty of authoring Megakernels at scale, we turned to augmenting Torch Compile (aka Inductor). Inductor traces PyTorch code to produce an intermediate representation called the FX Graph. From there, we lowered all captured kernels to PTX, inserted optimizations like Programmatic Dependent Launch (PDL), and replaced with optimized kernels. Then, we recompiled the resultant graph into a single binary. Since the FX Graph cannot represent kernels launched from C++, we leveraged low-level tracing (e.g., CUPTI for NVIDIA, rocProf for AMD~\cite{ramos2026kerncap}) to augment the graph with the missing kernel launch information.
    \item \textbf{Expert-managed agent context.} We provided expert-curated knowledge bases and optimization strategies to seed each agent's optimization approach. We asked agents to log their research findings in a custom log data structure.
\end{enumerate}
As coding harness capabilities have improved (Claude Code, \texttt{/goal}, \texttt{auto mode}, better compaction), we found ourselves \textit{removing intermediate representations} to unlock higher performance ceilings. {\shortname{}} now directly provides the agent with the inference engine's source code and no explicit context management structures or curated strategies.

\paragraph{Traditional IRs have limited expressivity.} We hypothesized that IRs could help agents understand complex inference codebases, which contain multiple codepaths and kernel launch mechanisms.
When directly provided with the source code, we do observe that agents spend $\approx20\%$ of their time reading engine files to understand the relevant code paths and kernel calls within the engine code. However, we also find that unrestricted source code access unlocks a higher performance ceiling. Even our heavily endowed augmented Inductor IR was not expressive enough to permit certain optimizations. For example, using the source code representation to optimize Kimi K3, agents could rewrite the MLA metadata planning kernel, so long sequences could be split across multiple thread blocks during attention, which provided a $1.7-2.5\times$ speedup on the MLA kernel itself. Neither the default, nor the augmented FX graph would be able to mutate the host-side metadata.

\paragraph{Surprising findings on expert-curated context for agents.} 
A prior version of our system, as well as most agentic performance optimization systems~\cite{wang2025geak}, utilize a knowledge hub, require agents to use research and experimentation logs, or prescribe the order in which the agent should tackle optimizations. For instance, systems often rank kernels by estimating its roofline and potential wall-clock speedup. Over time, in light of improvements in coding agent abilities, we  ablated more of these choices and found surprising results:

\begin{itemize}
    \item \textit{Knowledge hub.} Originally, our production system used a  knowledge hub seeded by experts. Upon ablating, we found that this approach can lead to 25--31\% lower speedup than agent performance without a knowledge hub after 12 hours of optimization (\Cref{fig:search-kb}). 
    \item \textit{Expert strategies.} Our system also seeded each parallel agent with a different expert-written strategy (e.g. fusion focused, communication-computation overlap, Megakernels). Upon ablating, we  found that simply giving each agent a simple prompt for optimization (``make this forward pass as fast as possible") outperforms the expert prompts (\Cref{fig:search-heuristic}). 
    \item \textit{Research and logging.} We find that providing human-engineered data structures for the agents to log and track their research also impedes throughput. Agents spend up to 43\% of the CLI/tool calls per trial on research tracking and data structure inspection, time that could instead be spent on optimization.
\end{itemize}

\noindent Additional details on the expert-curated context are provided in Appendix \ref{app:addlexp}. Together, these experiments point towards a bitter lesson~\cite{sutton2019bitterlesson} in agentic performance optimization. However, while we find that prescriptive IRs can constrain agent optimization, structured representations can be valuable when verifying the correctness of compiled programs. Ideally, future work could design an IR for verifying agent-generated kernels, without restricting the optimization space.

%% file: Tables/tab_4_3_engine_startup_arxiv.tex
\begin{table}[t]
\centering
\small
\setlength{\tabcolsep}{4pt}
\renewcommand{\arraystretch}{1.15}

\begin{subtable}[t]{0.51\linewidth}
\centering
\begin{tabular}{@{}lrr@{}}
\toprule
Phase & Time (s) & Share (\%) \\
\midrule
Model loading               & 265 & 34.8 \\
KV-cache sizing             & 73  & 9.6  \\
Autotuning                  & 192 & 25.2 \\
CUDA-graph capture, decode  & 34  & 4.5  \\
CUDA-graph capture, prefill & 192 & 25.2 \\
\midrule
\textbf{Total time to serving} & \textbf{761} & \textbf{100.0} \\
\bottomrule
\end{tabular}
\caption{Engine cold-startup time}
\label{tab:m31flash-boot}
\end{subtable}%
\hfill
\begin{subtable}[t]{0.47\linewidth}
\centering
\begin{tabular}{@{}lrr@{}}
\toprule
Metric & Slow tier & Fast tier \\
\midrule
Layers served              & 60   & 12   \\
Startup to first job (min) & 11   & 3    \\
Hot reload (s)             & 45   & 17   \\
Job completion time, median (s)       & 64   & 28   \\
Measurement variance\ (\%)    & 0.14 & 0.05 \\
Crashes / 1,000 jobs       & 0    & 3.9  \\
\bottomrule
\end{tabular}
\caption{Tiered Persistent Engine statistics}
\label{tab:m31flash-tiers}
\end{subtable}

\caption{MiniMax-M3.1-Flash on TensorRT-LLM. We break down the startup time of an inference engine and showcase differences between tiers in the persistent engine.}
\label{tab:m31flash}
\end{table}

%% file: Figures/3_1_tiered_speedup.tex
\begin{figure*}[!t]
    \centering
    \includegraphics[width=\textwidth]{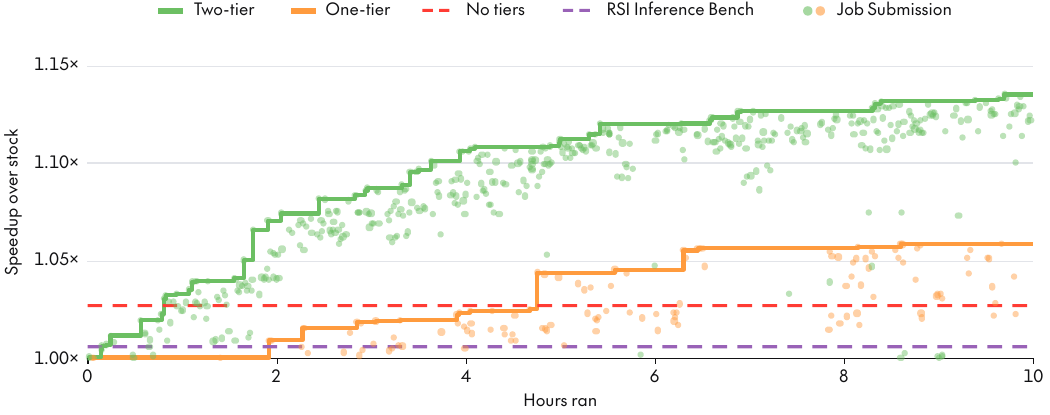}
    \caption{\shortname{}'s two-tiered measurement system outperforms a single slow tier, no measurement tiers, and Apple's RSI inference bench harness on B200 GLM 5.3 optimization.}
    \label{fig:tiered_speedup}
\end{figure*}

%% file: Sections/_patch_engine_code_snippet.tex
\begin{figure}[t]
\centering
\begin{lstlisting}[style=livepatch]
# Agent writes optimized forward pass in "tkcc_forward.py"
def forward(model, input_ids, positions):
    return optimized_forward(model, input_ids, positions)

# Agent submits job to server
post_rpc(engine, "hot_reload", file="tkcc_forward.py")
result = measure(engine)
post_rpc(engine, "teardown")

# Engine updates the forward function pointer
def hot_reload(file):
    write("/tmp/forward.py", file)
    candidate = exec_file("/tmp/forward.py")
    baseline = model.forward
    model.forward = candidate.forward       # patch
    recapture_cuda_graphs()
def teardown():
    restore(model, baseline)   # unload modules
    recapture_cuda_graphs()
\end{lstlisting}
\caption{Hot reloading function snippet. (1) The agent writes \texttt{tkcc\_forward.py}. (2) The agent sends it to the server over RPC. (3) Each rank writes the file to disk, runs \texttt{exec}, updates the model's forward pointer, recaptures CUDA graphs, and unloads kernel modules upon teardown.}
\label{fig:live-patch}
\end{figure}

%% file: Figures/cost_model_crossover/cost_model_crossover.tex
\begin{figure*}[t]
  \centering
  \begin{subfigure}[t]{0.48\textwidth}
    \centering
    \includegraphics[width=\linewidth]{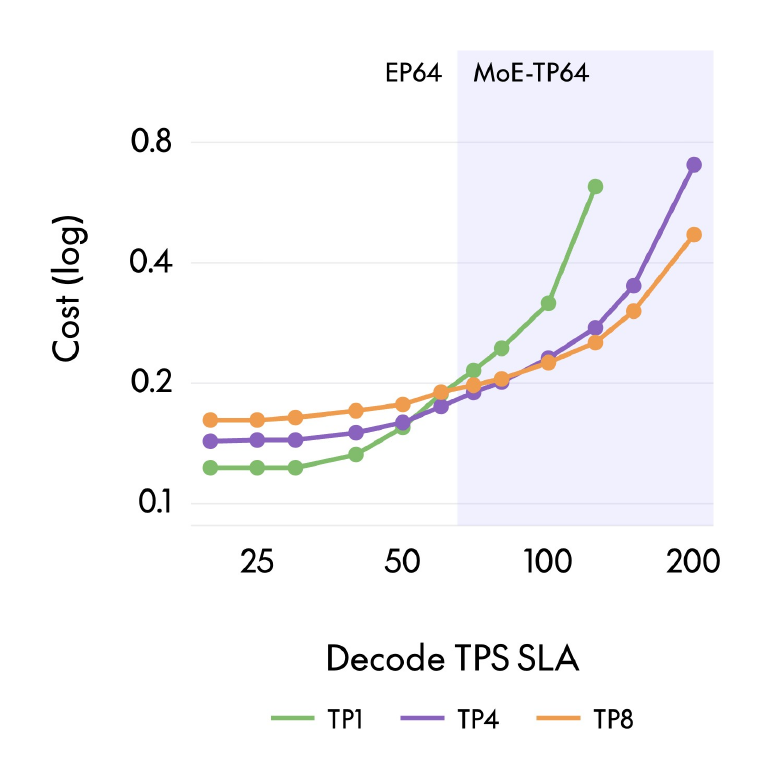}
    \caption{EP wins at higher concurrencies while higher TP wins at lower concurrencies.}
    \label{fig:k3-decode}
  \end{subfigure}\hfill
  \begin{subfigure}[t]{0.48\textwidth}
    \centering
    \includegraphics[width=\linewidth]{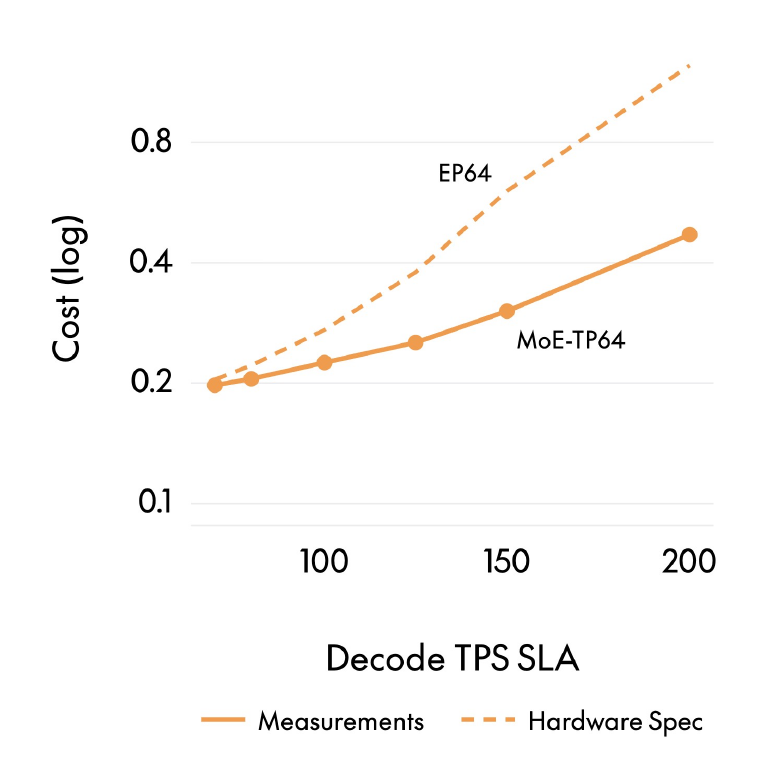}
    \caption{Hardware specs favor EP since it ignores per-call latency for collectives.}
    \label{fig:k3-specs}
  \end{subfigure}
  \caption{\textbf{Cost Model Crossover Points.}  As decode TPS increases, the best configuration uses TP instead of EP for MoE layers and higher degrees of TP for the attention and dense weights.}
  \label{fig:crossover}
\end{figure*}


%% file: Figures/3_3_search.tex
\begin{figure*}[!t]
  \captionsetup[subfigure]{labelformat=parens, justification=centering}
  \centering
  \begin{subfigure}[t]{0.32\textwidth}
    \centering
    \includegraphics[width=\linewidth]{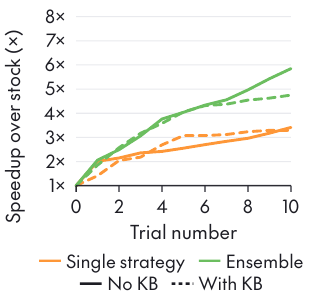}
    \caption{Inclusion of a knowledge base can incur 26\% worse performance.}\label{fig:search-kb}
  \end{subfigure}\hfill
  \begin{subfigure}[t]{0.32\textwidth}
    \centering
    \includegraphics[width=\linewidth]{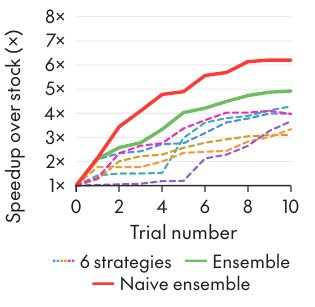}
    \caption{Expert-tuned ensembling is 1.3$\times$ slower than naive strategies.}\label{fig:search-heuristic}
  \end{subfigure}\hfill
  \begin{subfigure}[t]{0.32\textwidth}
    \centering
    \includegraphics[width=\linewidth]{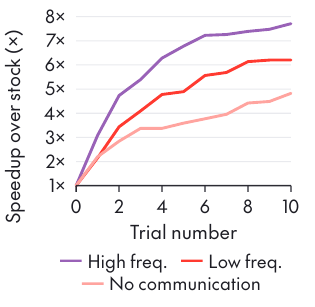}
    \caption{High-frequency agent communication is 4.7$\times$ faster.}
    \label{fig:search-naive}
  \end{subfigure}
  \caption{Ablating techniques for agent strategies during search on MiniMax-M3 layer 3, B200 TP4, batch size 1 with TensorRT-LLM.}\label{fig:search_strats}
\end{figure*}

%% file: Sections/5_results.tex
\label{sec:sec5_results}
\input{Figures/5_1_llama_results}

We evaluate \name{} on a variety of workloads against the strongest available open-source baselines. We find that {\shortname{}}:
\begin{enumerate}
    \item \textbf{Outperforms hand-tuned Megakernels.}  \name{} outperforms highly-optimized Megakernels (ThunderKittens~\cite{spector2025look} and Mirage Persistent Kernel~\cite{cheng2026mpk}) on their workhorse Llama 1B workload by up to $1.7\times$ (TPS) on batch size 1. We evaluate on H100 and B200 GPUs on output sequence lengths up to 64K tokens.
    \item \textbf{Tops engine leaderboards.}  \name{} adapts to different models, engine configurations, and hardware platforms, outperforming the best open performance on SemiAnalysis' InferenceX benchmark ~\cite{semianalysis2026inferencex} across frontier models (Moonshot Kimi K3~\cite{kimik32026}, Z.ai GLM 5.3~\cite{zai2026glm53}, DeepSeek v4.1 Flash~\cite{deepseekai2026deepseekv41flash}, and MiniMax M3.1 Flash~\cite{minimaxm32036}) by up to $2.6\times$ (TPS).

    \shortname{} also sets the state-of-the-art performance at the low-latency batch size 1 regime, generating 1111 TPS on B200 MiniMax 3.1 Flash and 1205 TPS on GB300 DeepSeek v4.1 Flash.
    \item \textbf{Generalizes across codebases and domains.} Third, \name{} generalizes beyond frontier inference, providing speedups of $1.5-6.4\times$, outperforming Claude Code on a variety of models spanning weather, biology, time series forecasting, and partial differential equations (PDE) simulation.
\end{enumerate} 

\subsection{Optimizing traditional workhorse workloads}

Figure~\ref{fig:llama-tps} shows that \shortname{} achieves state-of-the-art performance on batch size 1 Llama 1B inference running on both H100 and B200s and across a range of output sequence lengths. This was achieved without the typical megakernel abstractions. Upon investigation, we find gains from the fine-grained memory paging system to be small (worth only 3.1\% of the full forward pass) and that driver-level primitives such as PDL, CUDA graphs, and L2 prefetching are able to recover the benefits of a single persistent kernel.
\input{Figures/5_2_tkcc_results_p90}
\input{Figures/5_2_tkcc_results_c1_p90}

\subsection{Topping inference leaderboards}

We evaluate the improvements that {\shortname{}} yields over the top AgentX configuration per concurrency, model, and hardware (as of October 27, 2026). 

\paragraph{Protocol. }For each setting, we use the same parallelism strategy that was used by leading configuration to isolate the effect of the agentic optimization system. For optimization compute, {\shortname{}} uses $2\times$ the replica size per model compilation; this supports our tiered persistent engine. We evaluate on B200, B300, GB300, MI355X, and Trainium2, demonstrating that the system generalizes across hardware platforms. We compile just once per parallelism configuration and use the same resulting forward pass for each concurrency that uses the same strategy. For AMD and NVIDIA, we use the AgentX traffic profile as the workload. We evaluate Trainium2 using a synthetic multi-turn dataset that uses $70$k token ISL, shorter contexts than AgentX, to focus on accelerator performance instead of CPU offloading overheads (details in Appendix~\ref{app:experimental-analysis}).

\paragraph{Results.} In Figure~\ref{fig:tkcc-results-p90}, we show that \shortname{} provides up to $2.6\times$ improvement in TPS across settings in just a 24-hour compilation window per setting. This improvement stems from \shortname{}'s high-throughput engine, which improves agent experiment throughput by 252$\times$ and reduces idling time per agent by $11\times$. Notably, \shortname{} sets the state-of-the-art across different engines (vLLM, SGLang) and hardware platforms (B200, B300, MI355X, Trainium2). In \Cref{fig:tkcc-c1-results-p90}, we identified points where our cost model suggests a different parallelism strategy than the baseline image at concurrency 1. For MiniMax M3.1 Flash, the top-performing B200 image on InferenceX follows a TP4 configuration. However, at batch size 1, decode is memory-bound. \shortname{} suggests a TP8 configuration, which halves each GPU's weight traffic. With TP8, \shortname{} reaches 1,111 TPS, 1.8$\times$ higher than the best published result (633). This configuration may not be intuitive. Running the same InferenceX image under TP8 achieves similar performance to the chosen TP4 result. 

\paragraph{Continual compilation as workloads shift.}  \shortname{} is able to identify different optimizations across concurrency regimes, adapting to bursty traffic behavior. On Kimi K3, optimizations such as a persistent MoE down-projection boosts performance by 2.8-5\% across concurrencies 22-78 but is 6.6\% slower at concurrency 8. \shortname{} recognizes this and selectively applies  optimizations based on the active batch size. 

While we optimize open-source baselines to help readers contextualize our results, we have also used {\shortname{}} to optimize for production traffic across our internal engine configurations. Notably, when replaying our production traffic, we found that not all InferenceX baselines scale. Measured on our internal benchmarks, the leading InferenceX MiniMax M3.1 Flash image hits a capacity ceiling at low requests per second per replica, and the tail TTFT degrades 10-40$\times$ while 17\% of requests fail.

\subsection{Generalizing across codebases and domains}
We evaluate {\shortname{}} on domains and codebases beyond language model inference, specifically models for weather, biology, time series forecasting, and PDE simulation.  
We find that {\shortname{}}'s design generalizes to these domains, producing faster models that translate to gains in simulation accuracy at test-time.

\paragraph{Workloads}
We optimize the AIFS-ENS 2.0 \cite{lang2025multiscalelossformulationlearning} and Aurora 1.5 \cite{bodnarFoundationModelEarth2025} weather models; the Boltz-2 \cite{Passaro2025.06.14.659707} and Evo 2 \cite{brixiGenomeModellingDesign2026} biology models (on top of Anthropic's and Evo 2's baselines \cite{scienceAcceleratingOpensourceBiomolecular}); the Chronos-2 \cite{ansari2025chronos2univariateuniversalforecasting} and Panda-72M \cite{lai2026panda} time series foundation models; and Walrus \cite{mccabe2025walruscrossdomainfoundationmodel} and Poseidon-T \cite{herde2024poseidon} PDE foundation models (2D only).  

\input{Tables/tab_sciml_speedups}

\paragraph{Results.}
{\shortname{}} speedups over the baseline forward passes range from $1.5\times$ to $6.4\times$ (Table~\ref{tab:sciml-independent-speedups}), while maintaining bit exactness to the original forward pass. We compare {\shortname{}} to the baseline of using the Claude Code \texttt{/goal} command in Appendix \ref{app:sciml} and Figure \ref{fig:sciml-measured-trajectories}). {\shortname{}} achieves performance improvements at a faster rate than Claude Code.
Speedups can unlock model accuracy gains for tasks like weather modeling that require taking many samples to reduce statistical noise or models that benefit from higher-resolution processing. For instance, AIFS and Aurora rely on ensembling multiple perturbed samples to produce forecasts for metrics like mean sea-level pressure and air temperature. The optimizations generated by \shortname{} allows users to take more samples and reduce statistical noise (Table \ref{tab:weather-ensemble-quality}).

\input{Tables/tab_5_3_sciml_sampling}

%% file: Figures/5_1_llama_results.tex
\begin{figure*}[!t]
    \centering
    \includegraphics[width=0.9\textwidth]{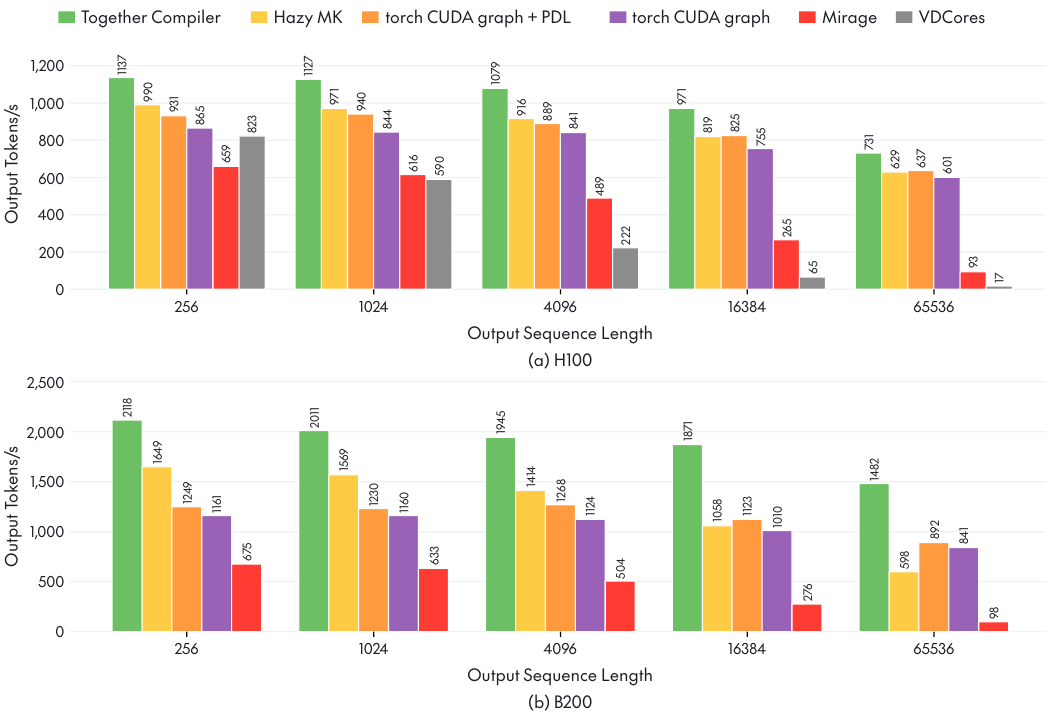}
    \caption{\shortname{} outperforms existing systems on Llama-3.2-1B-Instruct, Batch Size 1, ISL 32}
    \label{fig:llama-tps}
\end{figure*}

%% file: Figures/5_2_tkcc_results_p90.tex
\begin{figure*}[!t]
    \centering
    \includegraphics[width=0.95\textwidth]{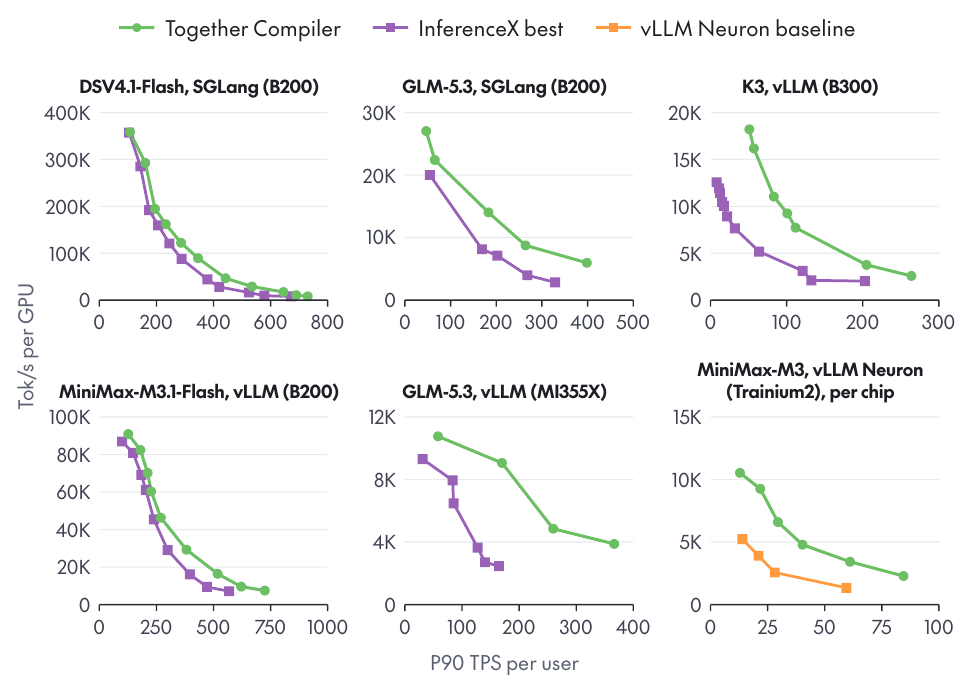}

    \caption{\shortname{} improves highly-optimized open-source baselines in under 24 hours of compilation, using at most 2 nodes per model. For these runs, we re-purpose the parallelism strategy proposed by the prior best InferenceX engine baseline as a starting point for our runs. The workloads and images used for these experiments are in Appendix \ref{app:experimental-analysis}.}
    \label{fig:tkcc-results-p90}
\end{figure*}

%% file: Figures/5_2_tkcc_results_c1_p90.tex
\begin{figure*}[!t]
    \centering
    \includegraphics[width=0.85\textwidth]{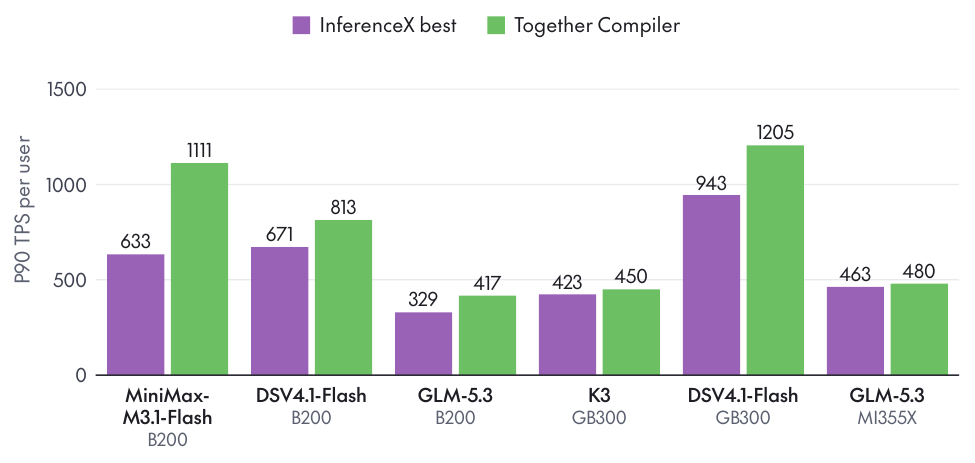}
    \caption{For these runs, we use our cost model to select the parallelism configurations that \shortname{}'s optimization loop focuses on, enabling state-of-the-art TPS in the low concurrency regime in under 24 hours of optimization. 
    }
    \label{fig:tkcc-c1-results-p90}
\end{figure*}

%% file: Tables/tab_sciml_speedups.tex
\begin{table}[t]
    \centering
    \setlength{\tabcolsep}{4pt}
    \renewcommand{\arraystretch}{1.15}
    \begin{tabular}{@{}ll lrr@{}}
        \toprule
        Model & Parameters & Evaluation & \textbf{TKCC} & \textbf{Claude Code} \\
        \midrule
        AIFS-ENS 2.0 & 230M & 36-case forecast eval & $3.5\times$ & Invalid \\
        Aurora 1.5 & 1.3B & ERA5: 54 cases & $3.7\times$ & $3.0\times$ \\
        Boltz-2 (Anthropic) & 521M & 7-size residue eval & $1.5\times$ & $1.2\times$ \\
        Evo 2 (Anthropic) & 7B & 16,000-token generation & $1.6\times$ & $1.6\times$ \\
        Chronos-2 & 120M & Chronos: 27 datasets & $2.6\times$ & $2.2\times$ \\
        Panda-72M & 72M & Chaotic ODEs & $1.9\times$ & $1.9\times$ \\
        Walrus & 1.3B & The Well: radiative layer & $4.3\times$ & $3.2\times$ \\
        Poseidon-T & 21M & NS-PwC & $6.4\times$ & $4.4\times$ \\
        \bottomrule
    \end{tabular}
    \caption{\shortname{} and Claude Code (\texttt{/goal}) speedups over baseline implementations on the respective evaluation datasets for each model. On AIFS-ENS 2.0, Claude Code fails to find a correct solution.}
    \label{tab:sciml-independent-speedups}
\end{table}

%% file: Tables/tab_5_3_sciml_sampling.tex
\begin{table}[t]
    \centering
    \setlength{\tabcolsep}{4pt}
    \renewcommand{\arraystretch}{1.15}
  \begin{tabular}{@{}llccc@{}}
        \toprule
        Model & Run & Samples & Temperature RMSE $\downarrow$ & CRPS $\downarrow$ \\
        \midrule
        AIFS   & Baseline & 4 & 0.97 K & 0.50 \\
               & \textbf{\shortname{}} & \textbf{13} & \textbf{0.94 K} & \textbf{0.46} \\
        \midrule
        Aurora & Baseline & 4 & 0.60 K & 0.25 \\
               & \textbf{\shortname{}} & \textbf{14} & \textbf{0.55 K} & \textbf{0.21} \\
        \bottomrule
    \end{tabular}
    \caption{\shortname{} enables more sampling in a fixed wall-clock time than the baseline implementation, corresponding to reductions in Root Mean Squared Error (RMSE) and Continuous Ranked Probability Score (CRPS) for temperature measurements.}
    \label{tab:weather-ensemble-quality}
\end{table}


%% file: Sections/6_conclusion.tex

The era of static inference is over. With evolving model architectures and customer use cases, we observe that inference has taken on many forms---it has now become ``polymorphic." The same optimizations no longer translate across the board, and each workload demands specialized deployments. With \name{}, we took agent-drive optimization from a prototype to a scalable production system for inference optimization. Our work shows that a successful agentic optimization loop can be built around three guiding principles: (1) Balance the measurement throughput-fidelity tradeoff; we accomplish this through a tiered persistent engine. (2) Establish the upper bound; we achieve this through a cost model that can reason through software immaturity. (3) Minimize human intervention; we deleted our old abstractions (custom IRs, knowledge bases, expert strategies) and left the agents in charge. The result is \name{}: an efficient system that boosts performance over highly-optimized baselines on the latest model releases by up to 2.6$\times$ in under 24 hours using just 2 nodes per model, while generalizing across hardware platforms and domains. For agent-drive optimization to be useful, it must be \textit{continual}.

%% file: Sections/appendix/7_appendix.tex
\noindent The appendix is organized as follows:

\begin{enumerate}[itemsep=0.6em]
    
    \item Appendix~\ref{app:experimental-analysis} provides further information on gathered experiments.
    
    \item Appendix~\ref{app:sciml} describes the results on scientific ML models in more detail.

    \item Appendix~\ref{app:addlexp} provides additional analysis and ablations for results described in the main paper.

    \item Appendix~\ref{app:cost-model} details the cost model and parallelism search space.

\end{enumerate}

\clearpage

\input{Sections/appendix/01_appendix_experimental_protocol}

\clearpage
\input{Sections/appendix/02_appendix_sciml}

\clearpage
\input{Sections/appendix/04_appendix_ablations}

\clearpage
\input{Sections/appendix/03_appendix_cost_model}
\clearpage
\input{Sections/appendix/05_appendix_agent_traces}

%% file: Sections/appendix/01_appendix_experimental_protocol.tex
\section{Experiment protocol}
\label{app:experimental-analysis}

This section provides more detail on how we ran the baseline systems and how we produced our main results.

\subsection{Baseline system protocols}

We discuss the protocol for running experiments on existing agentic performance optimization systems. For these experiments, we focus on the throughput of the \shortname{} optimizer and fix the baseline parallelism configuration. The coding harness for all systems was Claude Code with Opus 5.5, reasoning high. The edit surface is kernel and forward pass changes without allowing any quantization. The base images are the same as \shortname{} and listed in Appendix~\ref{app:experimental-analysis} (GLM 5.3 SGLang on B200 and GLM 5.3 vLLM on MI355X). The evaluation workload was the same AgentX workload, and the signal given to the agent was decode TPS.

\paragraph{RSI-Inference Bench.} We ran the default harness used by Apple's RSI-Inference Bench~\cite{liu2026rsiinference} (upstream commit \texttt{1a38072}) given a 10-hour budget and 8-hour verifier timeout. The base image is the same as the GLM 5.3 image used by \shortname{} (listed in Appendix~\ref{app:experimental-analysis}). We provide the RSI-Inference agent with the same edit surface as \shortname{} (i.e., the model's forward pass without any quantization allowed). We also use the same AgentX workload for evaluation and measure on output TPS. The result is provided in Figure~\ref{fig:tiered_speedup}.


\subsection{Megakernel baseline protocols}
\label{app:megakernels}

We sweep output sequence length at ISL 32 and batch size $1$. We compare to the following baselines on H100 and B200:
\begin{itemize}
    \item We compare to megakernels across this repository: \url{https://github.com/HazyResearch/Megakernels}
    \item We compare to Mirage using this repository: \url{https://github.com/mirage-project/mirage}
    \item We compare to VDCores using this repository: \url{https://github.com/vdcores/vdcores}
\end{itemize}

\subsection{InferenceX engine protocols}

We ran \shortname{} on four different model architectures across five hardware platforms, starting from their best configuration at the time of publishing. Optimization loops ran for 24 hours of continuous optimization and ran on 2$\times$ the replica size (one for the fast persistent engine and one for the slower persistent engine). Here we describe the datasets and protocols used for each setting. 

\paragraph{AgentX experiments.}
For the NVIDIA and AMD experiments, we evaluate end-to-end serving on SemiAnalysis InferenceX's AgentX benchmark and keep its methodology unchanged. That includes its pinned load generator (agentx-harness/AIPerf~1.0.12), benchmark command, public trace dataset (393 agentic-coding traces, seed~42), measurement window and scoring. We report InferenceX's metrics:


\begin{itemize}
  \item \textbf{Throughput per GPU:} input tokens (cached tokens counted in full) plus output tokens, divided by
        wall-clock time and by the number of serving GPUs.
  \item \textbf{Per-user interactivity:} $1/\text{P90}$ inter-token latency, and $1/\text{median}$ inter-token
        latency.
\end{itemize}

For \Cref{fig:tkcc-results-p90} and \Cref{fig:tkcc-c1-results-p90}, the baseline performance is sourced from the best InferenceX numbers at the time of writing. 
To check our protocol, we rerun it on InferenceX's published engine configurations and locally confirmed that it reproduces the published throughput exactly, and P90 interactivity to within about 2\%. We include the P50 results in \Cref{fig:tkcc-results-p50} and \Cref{fig:tkcc-c1-results-p50}.

\paragraph{Trainium Evaluation.} We optimize for a similar multi-turn synthetic dataset when evaluating Trainium. 
\begin{itemize}
    \item Prompts are built from public data: about 50\% Python code (codeparrot-clean), about 30\% Wikipedia prose (wikitext-103), and about 20\% generated JSON/YAML/Markdown. They are tokenized with M3’s tokenizer, seed 20261007.
    \item 65,792-token context with a 3,456 token system prompt shared across all users. Each request sees roughly 95\% cache hit rate. Output sequence length is exactly 700 tokens. Sampling parameters are temperature 1.0, top-p 0.95. Each user sends its next turn as soon as the previous turn finishes.
\end{itemize}

\paragraph{Configurations.} We report the engine configuration and images for each hardware platform below:

\input{Tables/tab_experiment_details}

\noindent The baseline inference images used to start each compilation are as follows:
\begin{itemize}
    \item \textbf{DeepSeekv4.1-Flash (SGLang, B200):} \texttt{\seqsplit{lmsysorg/sglang:nightly-dev-cu13-20261021-c1c287a4@sha256:b0473834763b2753e069cb96c1acaf235ba7fd632ccd9bbeb44da809c437fdeb}}
    \item \textbf{GLM-5.3 (SGLang, B200):} \texttt{\seqsplit{lmsysorg/sglang:v0.5.22-cu130@sha256:e3a4f06001a29542fc4a8638536d9ffc5eb9c9942b08603c09b19a56eae4a1ae}}
    \item \textbf{Kimi K3 (vLLM, B300):} \texttt{\seqsplit{vllm/vllm-openai:nightly-dev-x86\_64-cu13-3696c77}}
    \item \textbf{Kimi K3 (vLLM, GB300):} \texttt{\seqsplit{vllm/vllm-openai:nightly-e4a5b5537344b103ed10de7764378ef09604ef3d@sha256:6b90105076216e861cd30fae8828efef9263a0399137b894289f3b6a685bf4ff}}
    \item \textbf{MiniMax-M3.1-Flash (vLLM, B200):} \texttt{\seqsplit{vllm/vllm-openai:minimax-m31-flash@sha256:86f582dae25ea92802353727164ae4eb03b37b18baf9119b805ffaec2a7ef07c}}
    \item \textbf{GLM-5.3 (vLLM, MI355X):} \texttt{\seqsplit{vllm/vllm-openai-rocm:nightly-rocm100-29468dde8b515031dc6d4d9d06bf0a2fa0442098}}
    \item \textbf{Minimax-M3 (vLLM-neuron, Trainium2):} PyTorch \texttt{\seqsplit{2.11.0}}, vLLM version \texttt{\seqsplit{0.24.0}}, vLLM-neuron upstream release \texttt{\seqsplit{upstream release-0.24.0.1.1.0}}, neuron-cc compiler commit \texttt{\seqsplit{2.27.5334.0+f702b353}}, libtorch-neuronx-lite \texttt{\seqsplit{2.11.0.1.0.1284+f49d8626}}, aws-neuronx-runtime-lib	\texttt{\seqsplit{2.34.10.0}}
\end{itemize}

\newpage
\input{Figures/5_2_tkcc_results_p50}
\input{Figures/5_2_tkcc_results_c1_p50}

%% file: Tables/tab_experiment_details.tex
\begin{table}[H]
\centering
\scriptsize
\setlength{\tabcolsep}{3pt}
\begin{tabular*}{\linewidth}{@{\extracolsep{\fill}}llllllllll@{}}
\toprule
 & & & & & \multicolumn{2}{c}{Precision} & \multicolumn{3}{c}{Parallelism} \\
\cmidrule(lr){6-7} \cmidrule(l){8-10}
Model & Engine & Hardware & Chips & Conc. & Base & Experts & DP & Attention & MoE \\
\midrule
DeepSeekv4.1-Flash & SGLang & B200 & --- & 1--256 & FP4 & FP4 & \{1, 4\}
  & \makecell[tl]{\{TP4,\\ TP4 + DPA\}} & \{TP4, EP4\} \\
GLM-5.3 & SGLang & B200 & --- & 1--48 & FP4 & FP4 & \{1, 8\} & \{TP4, DPA\} & TP4 \\
Kimi K3 & vLLM & B300 & --- & 1--70 & FP4 & FP4 & --- & TP, DCP8 & TP \\
MiniMax-M3.1-Flash & vLLM & B200 & --- & 1--56 & FP4 & FP4 & --- & TP4 & TP4 \\
GLM-5.3 & vLLM & MI355X & --- & 1--16 & --- & --- & --- & --- & --- \\
MiniMax-M3 & \makecell[tl]{vLLM\\ Neuron} & Trainium2 & \{4, 8, 16\} & 1--128
  & \makecell[tl]{MXFP4/\\FP8} & \makecell[tl]{\{MXFP4,\\ FP8\}} & \{1, 2, 4\}
  & \makecell[tl]{\{TP16,\\ TP16 + CP4\}}
  & \makecell[tl]{\{EP2$\times$TP8,\\ EP4$\times$TP8,\\ EP8$\times$TP8\}} \\
\bottomrule
\end{tabular*}
\caption{Engines, hardware, precision, and parallelism configurations used to seed \shortname{}.}
\label{tab:seed-images}
\end{table}

%% file: Figures/5_2_tkcc_results_p50.tex
\begin{figure*}[!t]
    \centering
    \includegraphics[width=1\textwidth]{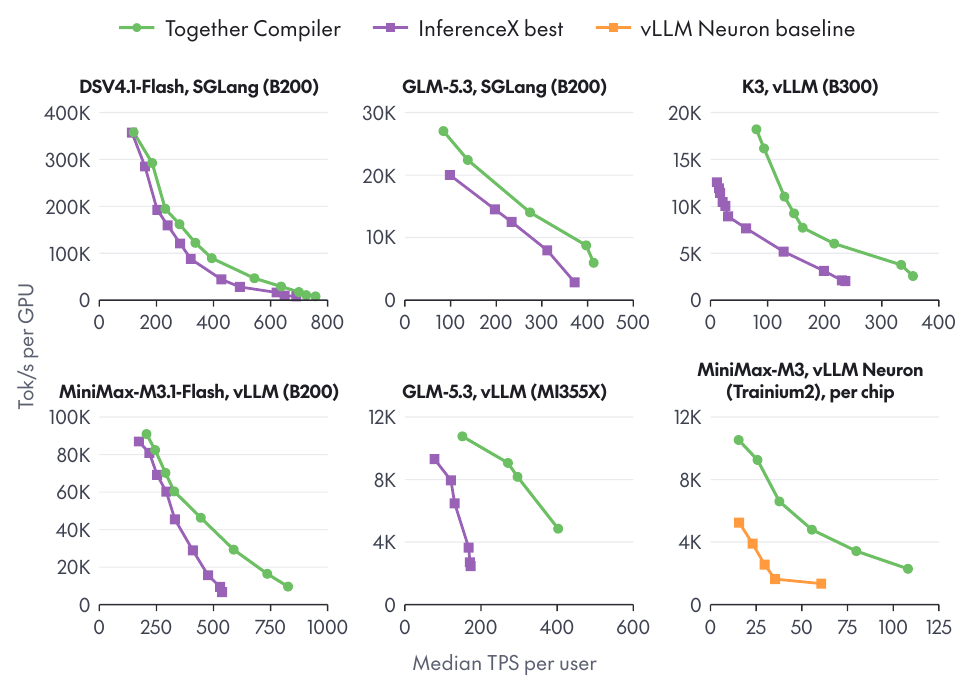}
    \caption{\shortname{} improves median TPS across production workloads.}
    \label{fig:tkcc-results-p50}
\end{figure*}

%% file: Figures/5_2_tkcc_results_c1_p50.tex
\begin{figure*}[!t]
    \centering
    \includegraphics[width=0.9\textwidth]{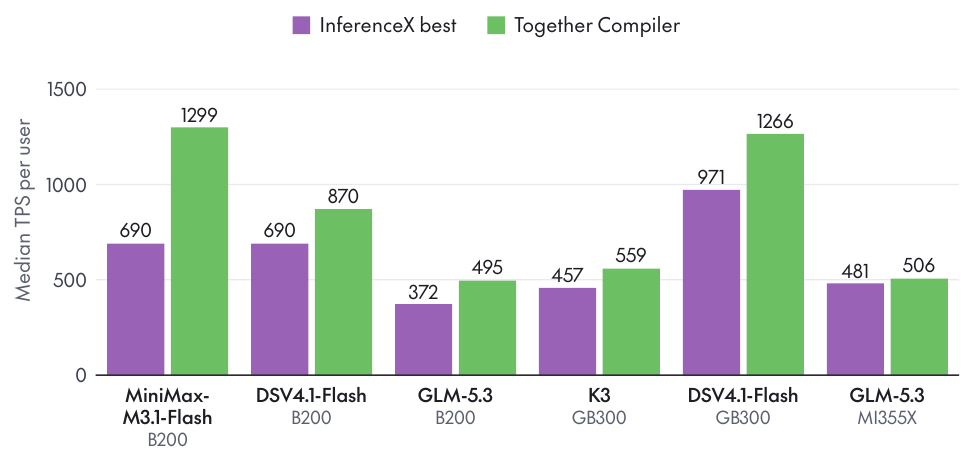}
    \caption{\shortname{} achieves over 1,299 TPS and 1,266 Median TPS on MiniMax-M3.1-Flash and DeepSeek-v4.1-Flash respectively.}
    \label{fig:tkcc-c1-results-p50}
\end{figure*}

%% file: Sections/appendix/02_appendix_sciml.tex
\section{Diverse Workloads}
\label{app:sciml}
We provide additional details on the set of diverse workloads we compare to a naive Claude Code baseline. For each, we adapted the model to standardize the weight loading, warmup, and candidate evaluation. Additionally, since multiple models relied on injected noise in their forward passes, we required bit exactness in outputs as well as in RNG states and other critical metadata.

\paragraph{Model Overviews}
\begin{itemize}
    \item \textbf{Aurora 1.5 Ensemble} is a weather foundation model that takes in atmospheric states, surface states, and forecasts future weather.  Repository: \url{https://github.com/microsoft/aurora/blob/b628d7ce93a8537e4844e41fc8aa2d44f522c674/aurora/model/aurora.py}.

    \item \textbf{AIFS-ENS 2.0} is a probabilistic weather model that takes in two global weather states six hours apart, along with static and time-dependent forcing fields. It predicts the state six hours after the latest input. Repository: \url{https://github.com/ecmwf/anemoi-core/blob/models-0.11.2/models/src/anemoi/models/models/ens_encoder_processor_decoder.py}.


    \item \textbf{Boltz-2} is a biomolecular model that predicts 3D structures of proteins, nucleic acids, and small molecules.  Repository: \url{https://github.com/anthropics/uplifting-biomolecular-modeling/tree/f4f62fa6592ae4938d49b1757bea0cfeff9f468e/boltz2/opt}.


    \item \textbf{Evo 2} is an autoregressive model that predicts nucleotides in DNA sequences. It follows a StripedHyena2 backbone. Repository: \url{https://github.com/anthropics/uplifting-biomolecular-modeling/tree/f4f62fa6592ae4938d49b1757bea0cfeff9f468e/evo2/opt}.


    \item \textbf{Chronos-2} is a T5 based, encoder only, time series foundation model that takes in a multivariate time series as input and outputs their forecasts. Repository: \url{https://github.com/amazon-science/chronos-forecasting/blob/1224fafa91e1ecd6674f5204066b159e40b5e4a1/src/chronos/chronos2/model.py}.


    \item \textbf{Panda-72M} time series foundation model is a PatchTST based, encoder-only transformer model.  Repository: \url{https://github.com/abao1999/panda/blob/c229e7c8c49cbe294458c44160248bb17856a715/panda/patchtst/patchtst.py}.

    \item \textbf{Walrus} is a PDE foundation model that simulates the dynamics of fluid-like physical systems. 
    Repository: \url{https://github.com/PolymathicAI/walrus/blob/73f2cf6b4c5871c8f5434802725b55666c66b5ab/walrus/models/isotropic_model.py}.

    \item \textbf{Poseidon-T}, unlike Walrus, takes in a single input state and an additional time step, and outputs the state after that amount of time. 
    Repository:
    \url{https://github.com/camlab-ethz/poseidon/blob/b8fa28f59bd7f7673323f28d11a12c6f3a215c61/scOT/model.py}.
\end{itemize}

\paragraph{Evals} Each model was tested for bit-exactness on an H100. The full list of evals can be found in Table \ref{tab:sciml-bit-exactness}.

\paragraph{TKCC vs Claude Code} Our Claude Code baseline used very similar context to the TKCC setup, where we gave it the same baseline references along with detailed descriptions of our evals and correctness requirements. We didn't, however, give it access to a persistent engine, a standard model adapter, and correctness tests--all of which we expected Claude to implement itself if it found them useful. Our choice of baseline was deliberately chosen so as to provide insights into what advantages the TKCC design provided over pure prompting on Claude Code.
    

\newpage
    
\input{Tables/tab_sciml_eval_coverage}
\input{Figures/sciml/speedup-vs-time-measured}

%% file: Tables/tab_sciml_eval_coverage.tex
\begin{table*}[t]
            \centering
            \scriptsize
            \setlength{\tabcolsep}{2pt}
            \renewcommand{\arraystretch}{1.2}
            \begin{tabularx}{\textwidth}{
                @{}
                >{\raggedright\arraybackslash}p{0.13\textwidth}
                >{\raggedright\arraybackslash}X
                >{\raggedright\arraybackslash}p{0.12\textwidth}
                >{\raggedright\arraybackslash}p{0.10\textwidth}
                >{\raggedright\arraybackslash}p{0.15\textwidth}
                @{}
            }
                \toprule
                Model & Evaluation coverage & Reference & TKCC & Claude Code \\
                \midrule
        
                AIFS-ENS 2.0 &
                36 forecast cases: three initialization times, four members,
                two rollout states; 24 single-member and 12 paired-member cases.
                Matched Baseline-noise protocol. &
                Baseline &
                Passed &
                Failed \\
                \addlinespace
        
                Aurora 1.5 Ensemble &
                54 ERA5 forecast cases: six initializations, lead times of
                1--6 hours; 36 cases at batch size 1 and 18 at batch size 2.
                Matched per-member RNG. &
                Baseline &
                Passed &
                Passed \\
                \addlinespace
        
                Boltz-2 &
                Seven residue sizes, 200--1,400, using constructed
                barnase/barstar inputs with empty MSAs; 10 recycles,
                200 diffusion steps, and 5 samples. &
                Anthropic exact &
                Passed &
                Passed \\
                \addlinespace
        
                Evo 2 &
                Evo 2-7B generation from a 3,000-base rCRS prompt:
                16,000-token warmup and measured generation in each of
                three matched repetitions. &
                Anthropic exact &
                Passed &
                Passed \\
                \addlinespace

                Chronos-2 &
                Forward panel from 27 dataset configurations:
                357 strata representing 5,973 original calls.
                &
                Baseline &
                Passed &
                Passed \\
                \addlinespace
        
                Panda-72M &
                Chaotic-ODE forward panel: 154 strata representing
                4,628 original calls. &
                Baseline &
                Passed &
                Passed \\
                \addlinespace
        
                Walrus &
                The Well's \texttt{turbulent\_radiative\_layer\_2D}:
                representative inputs for two batch-size buckets,
                weighted by 1,184 original calls. &
                Baseline &
                Passed &
                Passed \\
                \addlinespace
        
                Poseidon-T &
                NS-PwC forward panel. Additional TKCC full-evaluator
                checks cover 8,640 samples, 240 trajectories, 135 batches,
                returned loss, and 39 non-timing metrics. &
                Baseline &
                Passed &
                Passed \\
        
                \bottomrule
            \end{tabularx}
            \caption{
                Verified bit-exact agreement with the designated reference
                on the tested workloads. Boltz-2 and Evo 2 use Anthropic's shipped exact baselines rather than
                unmodified Baseline.
            }
            \label{tab:sciml-bit-exactness}
            \vspace{-2mm}
\end{table*}

%% file: Figures/sciml/speedup-vs-time-measured.tex
\begin{figure*}[h]
    \centering
    \includegraphics[width=0.75\linewidth]{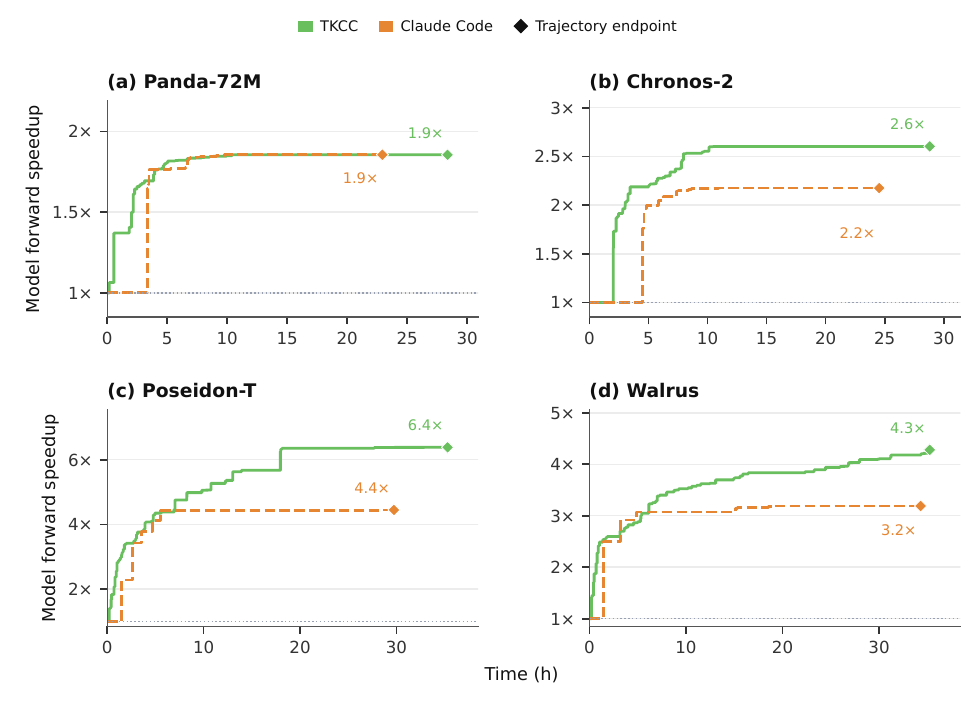}
    \vspace{-3mm}
    \caption{Best model-forward speedup found versus time for TKCC and Claude Code, with gaps between sessions removed, and final speedup of optimization marked by solid diamond. The significant time it took for Claude Code to obtain a candidate optimization is explained by the fact that it had to develop its own tooling/adapter for the model based off the context given to it by the codebase. TKCC and Claude Code performances are very similar for Panda, but we see a clear advantage in using TKCC over Claude Code for the other three models in terms of time and performance.}
    \label{fig:sciml-measured-trajectories}
\end{figure*}

%% file: Sections/appendix/04_appendix_ablations.tex
\section{Additional experiments and analysis}
\label{app:addlexp}

\subsection{Measurement fidelity for data-dependent architectural components}

Modern AI architectures use data-dependent execution paths governed by sparse top-$k$ attentions and routed mixture-of-experts.  Here we include examples of kernels generated based on random data distributions that may struggle to generalize to real data distributions during inference.

\paragraph{Sparse top-$k$ attention.}

Below we include an example of an Opus 5.5 agent-authored sparse top-$k$ attention kernel for GLM 5.3 on AMD MI355X. The kernel provided a speedup when benchmarked on random input data, but faces high run times when benchmarked on real data.

\begin{figure}[h!]
\centering
\begin{lstlisting}[style=livepatch]
// tk_topk5: exact top-K (K = 2048) of one row of indexer scores (excerpt, simplified).
// Source: tkcc champion kernel topk5.hip (GLM-5.3 layers 1-2, MI350X).

// 1) Bucket scores into 2048 histogram bins; find the threshold bin T holding the K-th largest score.
const uint32_t T  = cut[0];        // threshold bin
const int nAbove  = cut[1];        // scores in bins above T: all kept
const int cntT    = cut[2];        // scores tied in bin T (ReLU zeros: up to the whole context)
const int need    = K - nAbove;    // how many of bin T to keep

// 2) Rank every candidate in bin T against every other candidate: O(cntT^2).
if (cntT <= CAPG) {                // CAPG = 32768 staged candidates
  for (int c = c0 + w; c < c0 + tc; c += NT / 64) {         // each candidate (one per wave)
    const uint32_t kt = ckey[c];  const int it = cidx[c];
    int rank = 0;
    for (int j = lane; j < cntT; j += 64)                   // vs. all cntT candidates
      rank += (ckey[j] > kt) | ((ckey[j] == kt) & (cidx[j] < it));
    for (int o = 32; o > 0; o >>= 1) rank += __shfl_xor(rank, o);
    if (lane == 0 && rank < need) out[nAbove + rank] = it;
  }
} else {
  // 3) Overflow fallback: rescan the whole row per candidate: O(cntT * len).
  for (int i = beg + tid; i < end; i += NT) {
    if (hbin(lp[i]) != T) continue;
    const uint32_t kt = fkey(lp[i]);
    int rank = 0;
    for (int j = 0; j < len; ++j)                           // entire row (len = context length)
      if (hbin(lp[j]) == T)
        rank += (fkey(lp[j]) > kt) | ((fkey(lp[j]) == kt) & (j < i));
    if (rank < need) out[nAbove + rank] = i;
  }
}

\end{lstlisting}
\caption{A top-$k$ kernel written by the agent that is fast on random data but slow on real data. The kernel scores every earlier token in the context and must return the indices of the $k$ highest-scoring tokens. It groups the scores into histogram bins and keeps every bin above the one that contains the $k$-th largest score. Within that bin, it ranks each score by comparing it against every other score in the bin, breaking ties by lower token index, so the run time grows quadratically with the bin's size. Random test inputs spread across many bins, so the kernel took 20$\mu$s. Real scores contain many exact zeros from ReLU, which all fall into one bin, so the kernel took up to 62 ms at 70K context.}
\label{fig:sparse-topk}
\end{figure}

\paragraph{Mixture-of-experts.} Below we include our analysis of Cursor's production mixture-of-experts (MoE) megakernel, which is used to train their models~\cite{sul2026mok}. The reference repository at the time of this writing uses random inputs in the kernel benchmark harness. \footnote{\url{https://github.com/cursor/mixture-of-kittens}} We observe that on random inputs, the proposed megakernel works well on GB300, but when there is more expert imbalance and large batch sizes, it is more efficient to use the copy engine for dispatch communication (\Cref{fig:fidelity}).

\input{Figures/glm_measurements}

\subsection{Agent context management bitter lessons}

In \Cref{method:lessons}, we describe our ablations for context management. Here we include additional details on the expert-curated knowledge base and optimization strategies, and we provide analysis into how the agent behavior changes with versus without this expert-curated context. Authors of the expert context were involved in the authoring of FlashAttention, ThunderKittens, HipKittens, MegaKernels, and many production-grade kernels at Together AI.

 Notably, most existing agentic optimization systems for end-to-end inference optimization use a knowledge bases~\cite{wang2025geak, ye2026cakecompiler, li2026agentickernels}, prescribe specific cost models or strategies for agents to follow during the optimization process (e.g., rank by end-to-end expected gain, roofline model, wall-clock speed) ~\cite{wang2025geak, tvm2026tirx, li2026agentickernels}, or instruct the agents to use specific data structures for logging research and experimentation~\cite{decagon2026autoresearch}. 

We first describe the contents of our knowledge base, strategies, and data structures and then analyze why agents may have struggled to use that context effectively.

 \subsubsection{Contents of the context shaping}

\paragraph{Expert-curated knowledge base.} In \Cref{fig:search_strats} (Left) we ablate the use of a knowledge base. Surprisingly, we find that the knowledge base can sometimes hurts performance. We include the following artifacts in our knowledge base:

\begin{itemize}
  \item \textbf{Fused-block recipes:} Based on prior production optimizations for each model, hardware architecture, batch size and context length, we include a schedule of the custom kernels that make up one transformer layer. Each recipe comes with a decision log explaining which operations were fused or kept separate, and why. We include per-session histories of tuning attempts, measurements and outcomes that led to each recipe.
  \item \textbf{Production kernels:} Kernel source code organized by operation (GEMM, grouped/MoE GEMM, attention, normalization, RoPE/QKV, quantization, top-$k$, collectives, routing), with one variant per architecture (NVIDIA Hopper/Blackwell, AMD CDNA3/CDNA4). Each kernel has shape metadata, measured performance and notes on design choices.
  \item \textbf{Reference kernels:} Example kernels collected from open-source kernel libraries, recorded as pinned pointers to an upstream commit and path, plus an index of GEMM shape configurations.
  \item \textbf{Kernel techniques:} Short write-ups of reusable optimization and correctness techniques in categories such as compute (MMA tiling, tensor-core instructions), memory (TMA, swizzling, caching), scheduling, occupancy, launch overhead, numerics, correctness under CUDA graphs, and benchmarking methodology.
  \item \textbf{Fusion patterns:} Descriptions of common fusion regions across operations, e.g.\ attention/FFN prologues, GEMM epilogues, MoE dispatch/finalize, and GEMM or attention fused with collective communication.
  \item \textbf{Research papers:} Distilled summaries of kernel and systems papers, with the full text attached.
  \item \textbf{Vendor documentation:} Local copies of the CUDA programming guide, PTX ISA and HIP/ROCm reference docs.
  \item \textbf{Third-party code:} Pinned copies of open-source kernel DSLs, kernel libraries and inference engines, with survey notes on each.
\end{itemize}

 The knowledge base uses a graph structure. Every entry is a Markdown file with YAML metadata (architecture, data type, operation kind, bottleneck type) and typed links to related entries. Agents navigate from one entry to the next rather than querying a database. We found that with the inclusion of a knowledge base, agents often wasted time finding relevant kernels and ideas they could reuse, whereas without a knowledge base similar performing kernels or strategies were achieved independently in a shorter period of time (Figure~\ref{fig:search-kb}).

\paragraph{Expert-curated optimization strategies.} In \Cref{fig:search-heuristic} we ablate ensembling naïve Claude Code agents that are all instructed to make the forward pass as fast as possible against ensembling agents that are each assigned a distinct optimization objective. These distinct objectives include:

\begin{itemize}
  \item \textbf{Quick wins:} Rank candidate changes with a written cost model, (share of step $\times$ confidence) / time to a measured result. Work from routing and stream changes, to drop-in kernels, to structural rewrites, banking each win before the next and re-ranking after every new best result.

  \item \textbf{Fusion:} Take ownership of the dominant kernels. For instance, agents replace closed-source GEMMs that cannot be fused with agent-authored open-source kernels. Agents then fuse across boundaries that a graph compiler like Inductor cannot reach (since closed-source kernels are opaque to them). This includes fusing  GEMM prologues and epilogues, and the next layer's norm folded into the current layer's residual add. 

  \item \textbf{Reduction:} Cut the decode step's launch count using only code already held (authored kernels and shipped fused ops). Target the runs of short copy, cast, norm and quantize launches between GEMMs, then tune each remaining launch per batch size.

  \item \textbf{Communication/compute overlap:} Build a ledger of the exposed time of every collective (all-reduce, MoE all-to-all), own each collective and its neighbouring GEMM, and measure each collective's floor on the machine. Hide collectives using producer-epilogue sends, per-tile signalling, split-grid persistent kernels, reduce-scatter + norm + all-gather fusion, and cross-layer pipelining.

  \item \textbf{Megakernel:} Grow a single persistent kernel step by step, one correct submission at a time: first the non-attention/non-MoE ``shell'' of a layer, then across layer boundaries, then absorbing attention and MoE (including in-kernel cross-GPU dispatch), and finally the whole decode step in one launch. Emphasis is on deadlock-safe grid synchronization and CUDA-graph safety.

  \item \textbf{Cost Model:} Treat benchmark submissions as the scarce resource. First measure the instrument's resolvable effect size $E$, and build a wall-time ledger that includes idle time and unexplained (``dark'') time. Then try to refute each candidate before spending a submission on it: does the row exist, is it on the critical path, does the mechanism match? Submit only candidates whose estimated ceiling exceeds $2E$. Notably, this strategy is representative of the approach used in many existing agentic inference systems \cite{li2026agentickernels, wang2025geak}. As a reduced number of strategies can be validated in engine at a time, a system must decide which optimizations to prioritize.
\end{itemize}

We find that prescribing agents to fixed strategies often led to wasted agent cycles. Fixed strategies cannot represent a dynamically changing environment, and constraining a prescribed order ahead of time forces agents to explore less optimal paths. On AMD GLM 5.3, we saw the megakernel agent spend 24-hours, only to come up with $0.22\%$ faster performance. In \Cref{fig:search-heuristic}, we see that ensembling naive agents outperforms distinct expert strategies (6.2$\times$ vs.\ 4.9$\times$ after 10 trials), and that a single naive agent beats every prescribed strategy alone (4.8$\times$ vs.\ 3.1--4.5$\times$). 

\begin{figure*}[!t]
    \centering
    \includegraphics[
        angle=270,
        width=0.9\textwidth,
        trim=0 0 25mm 0,
        clip
    ]{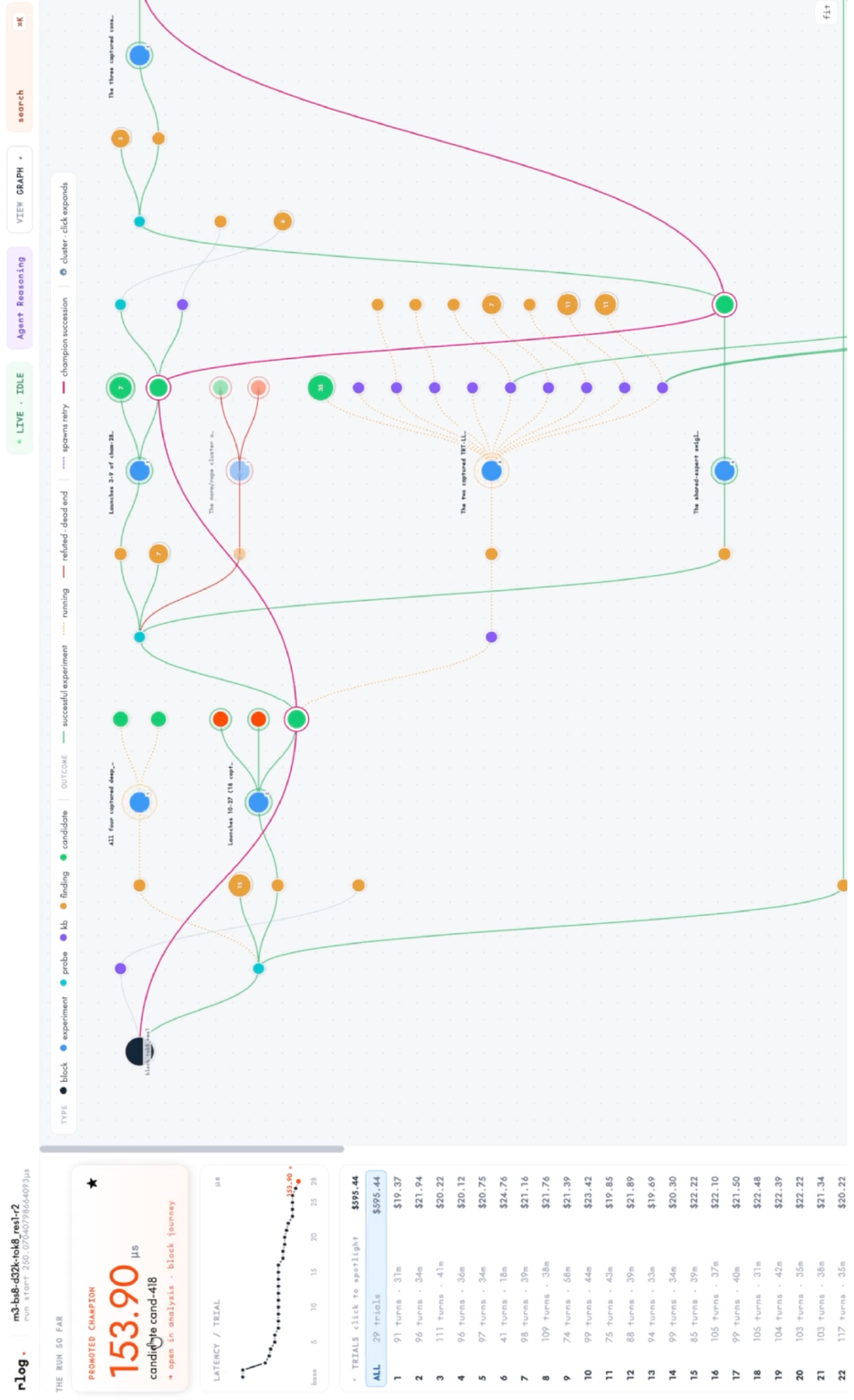}\par
    \vspace{-5mm}
    \includegraphics[
        angle=270,
        width=0.9\textwidth,
        trim=25mm 0 0 0,
        clip
    ]{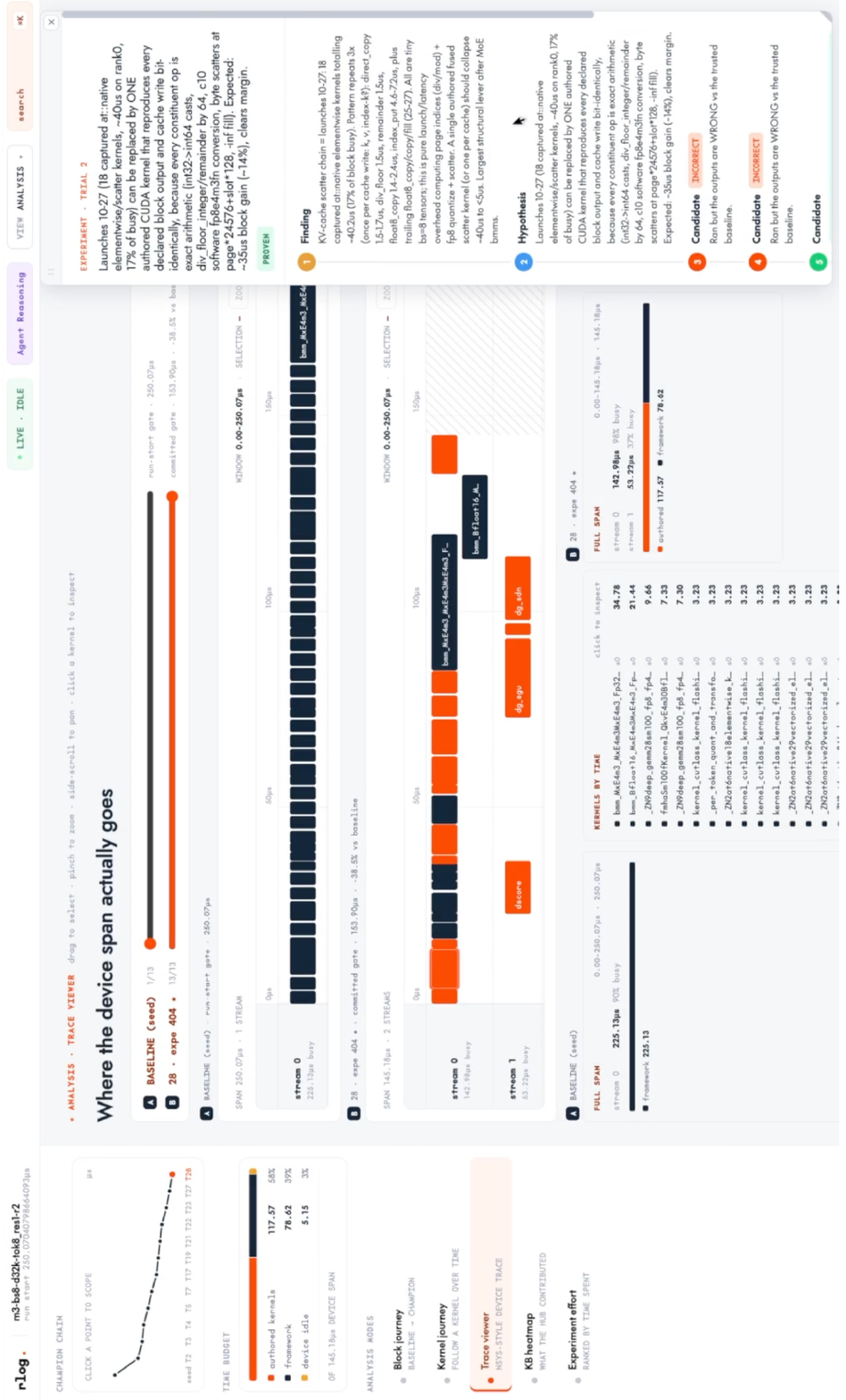}\par
    \vspace{-10mm}
    \caption{Examples of the experiment graph data structure that existed in earlier versions of {\name{}}. \textbf{Above}, we show nodes of the parallel agent experiment graph where blue dots represent new agent hypotheses, red dots indicate optimizations that led to slower performance, and green dots indicate optimizations that led to faster performance. The left panel shows a chart of the model's layer timing across optimization trials; the table below the chart shows the dollars spent per trial. \textbf{Below}, we show the trace viewer for the baseline kernel sequence and the {\shortname{}}-generated kernel sequence. On the right panel, we can see the hypotheses, findings, and experiments that the agents run.}
    \label{fig:graph-demo}
\end{figure*}

\paragraph{Data-structure for logging research and experimentation progress.}   \Cref{fig:graph-demo} shows examples of the graph data structure that we proposed to help agents keep track of experiment threads.

We persisted data using a knowledge graph that managed an agent's context at the end of each trial boundary. Nodes are typed: \textit{profile} nodes anchor measurements, \textit{finding} nodes record observations derived from profiles or accumulated history, \textit{experiment} nodes represent falsifiable hypotheses that close only when a measurement gate proves or disproves them, and \textit{candidate} nodes are agent-generated submissions for new optimizations. Edges represent lineage through the graph, as optimizations stack on top of each other over time. To query information from the graph, we provided the agent with a command line interface (CLI), ensuring that future trials can easily parse through previous iteration's findings.

When comparing end-to-end compiler runs that start from the same baseline inference engine on GLM 5.3 AMD and GLM 5.3 NVIDIA, we find that agents spend $8\%$ of time on bookkeeping, i.e.\ saving and looking up their past results in the graph through the CLI (${\sim}12.6$ seconds per call, since each call starts a new Python process), compared to $<0.5\%$ of time on bookkeeping when we run without our logging data structure and agents instead keep their notes in a plain Markdown file.

\subsection{Analysis of {\name{}} agent traces}
\label{app:agent-traces}

This section provides additional analysis into how agents spend time within the system. We discuss the positive actions that agents take within the system and the opportunities that we observe for further improvement.  

\paragraph{Throughput determines the ceiling of agent optimization.} {\name{}} improves throughput by 252$\times$ over the baseline RSI-Inference system that repeatedly relaunches the inference engine to test each new optimization idea. However, we find that \textit{throughput is still the primary outstanding bottleneck} to unlock further improvements in agent optimized inference systems. The {\shortname{}} measurement gates operate on the order of seconds, but still, we find that $>50\%$ of agent time can be spent waiting on feedback from the gates, waiting behind experiments queued by other agents, or agents re-sending the same code to re-check results when the code produced marginal improvements or slowdowns (\Cref{tab:tc-time-allocation}).

\begin{table}[h!]
  \centering
  \small
  \begin{tabular}{@{}lrrr@{}}
    \toprule
    Model (platform) & Waiting for gate (\%) & Thinking and writing code (\%) & Other (\%) \\
    \midrule
    GLM-5.3 (AMD MI350X)     & 57.0 & 36.9 & 6.1 \\
    GLM-5.3 (NVIDIA B200)    & 55.0 & 35.7 & 9.3 \\
    MiniMax M3.1 Flash (NVIDIA B200) & 52.7 & 40.4 & 6.9 \\
    \bottomrule
  \end{tabular}
  \caption{How {\name{}} agents spent their time. ``Other'' covers reading and editing files,
  compiling kernels, profiling and bookkeeping.}
  \label{tab:tc-time-allocation}
\end{table}

\paragraph{Optimizations discovered by hardware platform.} Here we discuss the key optimizations agents implemented. Some patterns appeared across hardware platforms and others were hardware-specific.

\noindent Common optimizations across platforms included:
\begin{itemize}
    \item \textbf{Fusing communication into compute.} For example, GLM 5.3 on AMD runs across $8$ GPUs, so every layer has several all-reduces, where the GPUs sum their partial results. The agents merged those into the kernels around them:
    \begin{itemize}
        \item One kernel does the all-reduce, the residual add, the RMSNorm and the FP8 quantization. The matrix multiply before it writes its output straight into the all-reduce's shared buffer, which removes a copy.
        \item The MoE down-projection is fused with the next all-reduce.
        \item The attention output projection is fused with its all-reduce and with part of the expert router.
    \end{itemize}
    On NVIDIA, agents in all four models wrote their own Lamport-style all-reduce, in which each GPU polls the data buffer for a sentinel value instead of waiting on a separate ready flag, and fused it with the residual add and RMSNorm (and, on MiniMax M3.1 Flash, the quantization). On one GLM 5.3 optimization run, Lamport and NVLS all-gather/reduce-scatter kernels took the score from $1.00\times$ (stock) to $1.26\times$ in the first $1.5$ hours, and on MiniMax M3.1 Flash the fused two-shot all-reduce ran about $2\times$ faster than NCCL at decode.

    \item \textbf{Deleting dead or wasted operations.} For example, with GLM 5.3 on AMD, agents found that three copy/memset operations in the captured graph did nothing. On NVIDIA, four DeepSeek V4.1 agents independently found that the stock FlashMLA decode kernel padded $16$ attention heads to $64$, doing four times the needed work. Removing the padding was the largest decode-attention gain in any DeepSeek V4.1 optimization run.
\end{itemize}

\noindent Common areas where agents invested a lot of time, but struggled to realize any performance improvements include:
\begin{itemize}
    \item \textbf{Modifying the attention kernel.} Agents spent significant effort on fusing neighboring kernels (e.g., norm, RoPE, output projection) into attention, or on changing how the work is split across GPU processors (e.g., split-KV). This applied both to the agents working on NVIDIA and AMD kernels.

    \item \textbf{Megakernel formation.} On GPUs, merging a layer, or long chains of dependent kernels, into one persistent kernel did not pay off. On AMD GLM 5.3 the result was only $0.22\%$ faster than the existing best version, and on NVIDIA it gave no lasting gain. Each dependent step still needed a grid-wide barrier, which cost $1.5$--$3.5\,\mu$s inside a kernel versus $0.5$--$1\,\mu$s at a kernel boundary with PDL. Trainium2 was the exception: its final builds run each decoder layer's attention and MoE as a single fused decode kernel.

    \item \textbf{Overlapping communication with computation.} On AMD, running RCCL all-reduces at the same time as GEMMs lost on every concurrency. On NVIDIA, related ideas (overlapping two batches, pushing the output projection into the all-reduce, split-K GEMMs) were each 3-4$\times$ slower across MiniMax M3.1 Flash runs.
\end{itemize}

\noindent Some of the hardware-specific optimizations and failure modes are as follows:
\begin{itemize}
    \item \textbf{NVIDIA:} On NVIDIA B200, B300 and GB300 GPUs, we optimized GLM 5.3, DeepSeek V4.1 Flash, Kimi K3 and MiniMax M3.1 Flash.
    \begin{itemize}
        \item \textbf{Programmatic dependent launch (PDL).} The next kernel launches early and loads its weights while the previous one finishes, waiting only where it needs the previous output. On DeepSeek V4.1, this cut the fixed cost of each decode step by $9.7\%$.
        \item \textbf{Reductions inside the NVLink switch.} NVIDIA's GPU-to-GPU switch can sum data itself (NVLS, \texttt{multimem}). This contributed to GLM 5.3's early step from $1.00\times$ to $1.26\times$ and to a Kimi K3 attention block that ran $1.68\times$ faster than stock. For the very small transfers in decode, though, it was slower than the agents' Lamport all-reduce. AMD MI350X GPUs connect to each other directly, without such a switch.
        \item \textbf{Blackwell low-precision tensor-core instructions.} The one agent-written matrix multiply that beat the vendor kernel used Blackwell's \texttt{tcgen05} instructions with NVFP4: a GLM 5.3 MoE down-projection that ran in $23.7\,\mu$s versus $34.1\,\mu$s for the vendor kernel.
        \item \textbf{Thread-block clusters.} Groups of thread blocks that run together and share fast local memory. Switching to an existing cluster-based attention kernel from the vendor library was the largest single improvement in one GLM 5.3 optimization run.
        \item \textbf{Tuning vendor libraries instead of replacing them.} Changing the tile sizes of DeepGEMM's MoE kernel cut its time from $273$ to $210\,\mu$s on DeepSeek V4.1. In contrast, agent-written replacements for vendor matrix multiply, attention and MoE kernels almost always lost, because the vendor kernels already ran at $78$--$100\%$ of memory bandwidth.
        \item \textbf{GPU resets.} Agent code with illegal memory accesses put GPUs into a state that required a full reset (driver error Xid 154): $9$ times during GLM 5.3, which led to $2$ machines being removed from the cluster.
    \end{itemize}
    \item \textbf{AMD:} On AMD, we optimized GLM 5.3 TP8 using FP8 precision weights.
    \begin{itemize}
        \item \textbf{ROCm scheduling tricks.} When a captured HIP graph splits into parallel streams, ROCm's scheduler decides which branch stays on the main stream. Agents found that recording the main computation first at each split keeps the critical path on the main stream.
        \item \textbf{Reusing AITER's all-reduce buffers.} AITER, AMD's kernel library, does its multi-GPU sums through special shared GPU-to-GPU buffers that bypass the cache. Agents had the preceding matrix multiply write its output directly into that buffer, which removes a copy and a kernel launch.
        \item \textbf{Filling gaps in AMD's tuning tables.} For several of GLM 5.3's matrix shapes, Composable Kernel (CK) had no tuned settings, so the code fell back to slow defaults. Agents added split-K variants for those shapes.
        \item \textbf{Compute-unit partitioning.} Attempts to reserve some compute units for communication while the rest ran computation (CU masks) were ignored on AMD.
        \item \textbf{Per-chip timers are unusable for timing.} MI350X/MI355X is built from 8 compute dies (XCDs), and each has its own clock. Agents found that reading the GPU timer inside kernels gave meaningless per-operation numbers. Thus, attempts to schedule work by die (``XCD remapping'') came out as ties.
    \end{itemize}
    \item \textbf{Trainium2:} On Trainium2, we optimized MiniMax-M3 serving long multi-turn conversations.
    \begin{itemize}
        \item \textbf{Fusing whole decoder layers into one kernel.} Custom NKI kernels run each decoder layer's attention and MoE as a single fused decode kernel. This is the opposite of what we saw on GPUs, where megakernels gave little or no gain. Other fused kernels cover the sparse-attention and MoE steps, the speculative-decoding verify and sampling step, and the KV-cache writes.
        \item \textbf{Choosing the parallel layout for each load level.} The split of work across chips (tensor, expert, context and data parallelism) was tuned separately for each number of concurrent users using our contributed cost model. For example, splitting the KV cache across chips helped from about $4$ users up but was slower at $1$--$2$ users.
        \item \textbf{Placing experts by measured routing.} At larger chip counts, experts are assigned to chips based on a measured profile of which experts tokens actually use, which balances the load across chips.
        \item \textbf{Choosing what to compile as one program.} At low load, the whole decode step is compiled as a single program; at high load, one layer is compiled and reused across layers.
    \end{itemize}
\end{itemize}

\paragraph{Closed model safety checks.} Several times, closed source models could stop optimizing altogether due to safety concerns. For instance, after two pieces of agent code crashed GPUs, agents were banned from editing the model’s recorded sequence of GPU steps directly. That stopped the crashes, but some of the improvements that had worked relied on exactly that technique, so they can’t be redone.

%% file: Figures/glm_measurements.tex
\begin{figure}[t]
  \captionsetup[subfigure]{labelformat=parens, justification=centering}
  \centering
  \begin{subfigure}[t]{0.32\linewidth}
    \centering
    \includegraphics[width=\linewidth]{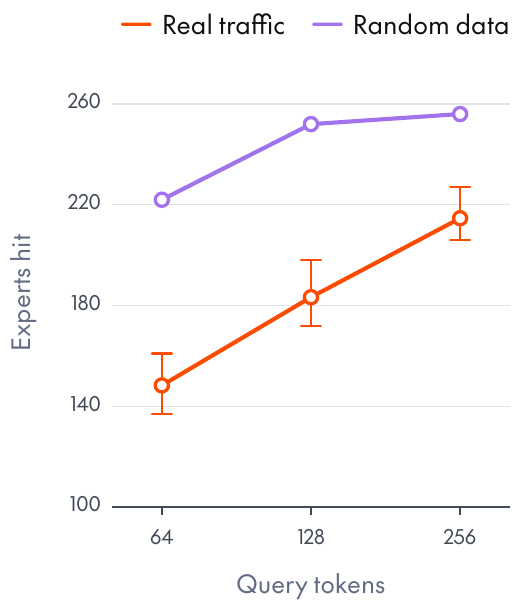}
    \caption{Random data activates more experts than real traffic.}\label{fig:fid-experts}
  \end{subfigure}\hfill
  \begin{subfigure}[t]{0.32\linewidth}
    \centering
    \includegraphics[width=\linewidth]{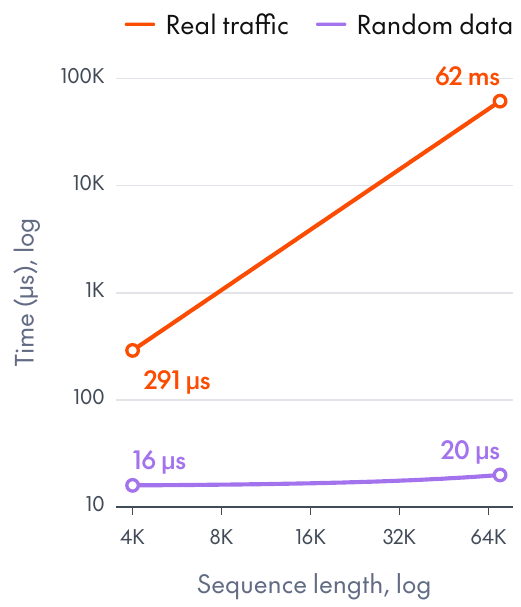}
    \caption{Top-$k$ kernel written assuming random data has quadratic slowdown.}\label{fig:fid-topk}
  \end{subfigure}\hfill
  \begin{subfigure}[t]{0.32\linewidth}
    \centering
    \includegraphics[width=\linewidth]{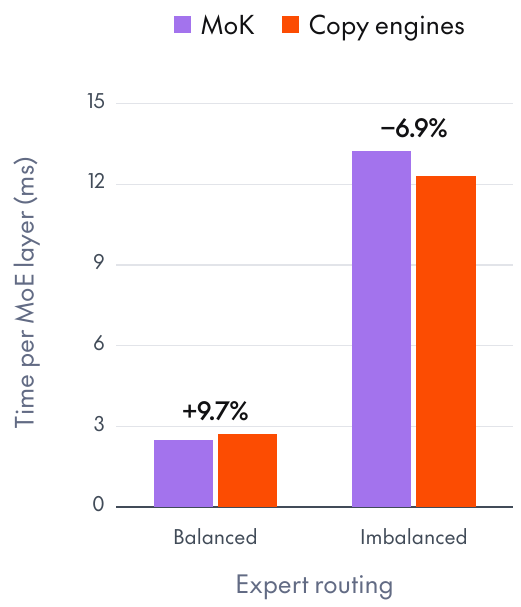}
    \caption{Copy engines beat MoK under imbalanced routing.}\label{fig:fid-copy-engine}
  \end{subfigure}
  \caption{Copy engine outperforms a production-grade MoE megakernel (MoK~\cite{sul2026mok}) in imbalanced routing.}\label{fig:fidelity}
\end{figure}

%% file: Sections/appendix/03_appendix_cost_model.tex
\section{Parallelism cost model}
\label{app:cost-model}
  One possible solution to the parallelism search problem is a full grid-search, where we enumerate a large number of
  possible parallelism schemes and run each scheme on GPU against a production-approximating workload. Theoretically,
  this approach would be of very high fidelity, given we actually collect timing information directly on the
  deployment for each parallelism configuration. However, grid-search has some notable downsides:
  \begin{enumerate}
      \item Results are dependent on software implementation for each parallelism strategy
      \item Time consuming, requiring engine startup and a benchmarking run for each strategy
  \end{enumerate}
  Therefore, we propose a cost-model-based approach to identify good parallelism candidates.

\paragraph{Notation} Symbols used throughout:
  \begin{center}
  \begin{tabular}{ll}
    \hline
    $N$, $N_{\max}$          & chips per replica; max replica upper bound (4096) \\
    $N_{dom}$                & chips in one scale-up domain \\
    $TP_{\text{attn}}, CP_{\text{act}}, CP_{\text{kv}}, DP$ & attention tensor, context (activations), context (KV cache), data parallel \\
    $TP_{\text{dense}}$      & dense tensor parallel \\
    $EP_{\text{MoE}}, TP_{\text{MoE}}$ & MoE expert parallel, MoE tensor parallel \\
    $\text{ISL}, \text{OSL}$ & input and output tokens per request \\
    $L$                      & layer index \\
    $T_\ast$, $S_\ast$, $F_\ast$ & a time, a byte count, a FLOP count \\
    \hline
  \end{tabular}
  \end{center}

  \paragraph{Admissible schemes}
  First we decide on the search space of admissible parallelism schemes. A scheme is admissible if all of the following
  rules hold:
  \begin{enumerate}
      \item Replica size
        \begin{enumerate}
          \item $N \leq N_{\max}$ \hfill (4096)
        \end{enumerate}
      \item MoE
        \begin{enumerate}
          \item Expert Parallel: $EP_{\text{MoE}}$ divides $E$, the number of routed experts, for every MoE block.
                $EP_{\text{MoE}}$ is not constrained by the scale-up-domain size.
          \item Tensor Parallel: $TP_{\text{MoE}}$ divides every expert's width where $N \leq N_{dom}$
        \end{enumerate}
      \item Dense
        \begin{enumerate}
          \item Tensor Parallel: $TP_{\text{dense}}$ divides $N$
        \end{enumerate}
      \item Attention
        \begin{enumerate}
          \item Tensor Parallel: $TP_{\text{attn}}$ divides every attention head and projection.
          \item Context Parallel (Activations): $TP_{\text{attn}} \cdot CP_{\text{act}}$ divides $TP_{\text{dense}}$
          \item Context Parallel (KV Cache): $CP_{\text{kv}}$ divides $TP_{\text{attn}}$ sharing one KV head
          \item Data Parallel: $TP_{\text{attn}} \cdot CP_{\text{act}}$ divides $N$, giving
                $DP = N / (TP_{\text{attn}} \cdot CP_{\text{act}})$
        \end{enumerate}
  \end{enumerate}

  \paragraph{Cost model}
  From there, we produced a cost model designed to estimate performance of a given model on a given chip and
  parallelism scheme. For each phase, the model computes the forward pass time from each model layer given parameters
  regarding compute FLOPs, HBM bandwidth, HBM capacity, and interconnect bandwidth, and it determines maximum concurrency
  allowed by memory capacity and latency target. From there, we can derive chip seconds per phase, which is then combined
  into one single throughput value: tokens/min per chip.

  \paragraph{Forward pass time}
  To calculate the forward pass time $T_{\text{pass}}$, let's define
  \begin{itemize}
      \item $T_c^L$, the time spent doing matrix compute at layer $L$
      \item $D^L$, the number of bytes --- both weights and KV cache --- loaded from HBM at layer $L$
      \item $BW_{\text{HBM}}$, the HBM memory bandwidth of the GPU
      \item $T^{L,i}_{\text{comm}}$, the time spent in GPU-to-GPU communication collective $i$ at layer $L$
  \end{itemize}
  Within a layer, compute and memory reads are assumed to overlap perfectly and max is taken per layer. Communication
  is assumed to be sequential with computation. For prefill, we assume it is compute bound and drop the memory load
  component.
  \begin{align}
      T_{\text{pass}}^{\text{decode}}  &= \sum_{L} \left[ \max\!\left(T_c^L,\ \frac{D^L}{BW_{\text{HBM}}}\right) + \sum_i T^{L,i}_{\text{comm}} \right], \\
      T_{\text{pass}}^{\text{prefill}} &= \sum_{L} \left[ T_c^L + \sum_i T^{L,i}_{\text{comm}} \right].
  \end{align}

  \paragraph{Concurrency}
  The concurrency $C$ is determined by the largest number of requests that satisfy two independent limits, given some
  fixed tokens/sec per user: 
  \begin{enumerate}
      \item Memory Limit $C_{\text{memory}}$: largest number of requests' KV cache that fit in HBM
      \item Speed Limit $C_{\text{speed}}$: largest number of requests whose pass still meets the latency target set by
            fixed tokens/sec per user
  \end{enumerate}
  \begin{equation}
      C = \min\left(C_{\text{memory}},\ C_{\text{speed}}\right)
  \end{equation}

  \paragraph{Disaggregated prefill and decode}
  Prefill and decode are modeled as disaggregated, where each phase runs on separate hardware with its own sharding
  scheme. For each phase, we evaluate every admissible scheme and keep top k lowest chip-seconds per request. Among
  schemes within $\epsilon$ of the best, we select based on fewest chips.
  \begin{align}
      \text{chip-s}_{\text{decode}}  &= \frac{N \cdot T_{\text{pass}}^{\text{decode}} \cdot \text{OSL}}{C},
      \\
      \text{chip-s}_{\text{prefill}} &= \frac{N \cdot T_{\text{pass}}^{\text{prefill}}}{C}.
  \end{align}

  \paragraph{Throughput}
  Finally, these are combined into a single measure of throughput, tokens/min per chip ($\text{TPM}$).
  \begin{equation}
      \text{TPM} = \frac{(\text{ISL} + \text{OSL}) \times 60}{\text{chip-s}_{\text{decode}} + \text{chip-s}_{\text{prefill}}}.
  \end{equation}

  \subsection{Attention}
  Attention is a key part of the model where there is a lot of variance from model to model. To properly account for
  all the variations, we characterize an attention layer with two variables: the family of an attention layer
  (based on the attention algorithm) and the key restrictions (decides which keys a query sees).

  \paragraph{Families}
  \begin{enumerate}
      \item \textbf{Softmax attention} scores each query against a set of keys ($QK^\top$) and sums their values ($PV$).
        \begin{enumerate}
          \item \emph{GQA}: $H$ query heads share $H_{kv}$ key/value heads, with head widths $d_{qk}$ (query/key) and
                $d_v$ (value). $H_{kv}$ changes what is cached and read, not the FLOPs: every query head still scores
                every key it sees.
          \item \emph{MLA}: the cache holds one latent of width $d_c$ plus a shared RoPE key of width $d_r$. Whether the
                up-projection is folded into the query (\emph{absorbed}) is set per phase: absorbed scores over
                $d_c + d_r$ and sums over $d_c$; unabsorbed scores over $d_{nope} + d_r$ and sums over $d_v$, where
                $d_{nope}$ is the per-head key width without RoPE.
        \end{enumerate}
      \item \textbf{Linear attention} (KDA) keeps a fixed $d_k \times d_v$ state per head instead of keys, so its work
            per token and its cache do not grow with context, and key restrictions do not apply.
  \end{enumerate}
  \paragraph{Key restrictions} A softmax layer may combine any of the following.
  \begin{enumerate}
      \item \emph{Window} $W$: a query sees at most the last $W$ tokens.
      \item \emph{Compression} $r$: stores one entry per $r$ consecutive kv cache tokens instead of one per token.
      \item \emph{Sparse selection} $s$: a scoring function ranks the kv cache entries and the query attends to the
            top $s$ of them (may be compressed or a block of tokens). The
            scorer is either a dedicated lightweight indexer or the attention scores from a coarser branch.
  \end{enumerate}
  A window combined with a kv cache (compressed and/or sparse) is \emph{additive}: the query attends to the
  union of its window and its selected kv cache entries in a single softmax or \emph{gated}: each branch is attended
  separately and the outputs are mixed by learned gates. 

  \paragraph{Query--key pairs} All softmax FLOPs follow from the number of query--key pairs. For a sequence with $q$ new
  tokens after $c$ cached ones, under a causal mask and a cap of $m$ keys per query where $m$ is set by the layer's key
  restrictions. The number of new tokens that see fewer than $m$ keys is
  \begin{equation}
    u = \max\!\big(0,\ \min(q,\ m - c - 1)\big),
  \end{equation}
  These $u$ tokens see $c + 1, \dots, c + u$ keys and the remaining $q - u$ see
  exactly $m$. This gives us $P$, the total number of query key pairs for one sequence in this pass:
  \begin{equation}
    P(q, c; m) = u\,c + \frac{u(u+1)}{2} + (q - u)\,m.
  \end{equation}

  Let $W$ be the window size, $r$ the compression ratio (tokens per kv cache entry; $r = 1$ without compression), and
  $s$ the number of kv cache entries kept by sparse selection. Since $s$ entries cover $s\,r$ tokens, the kv cache
  portion is counted in tokens with cap $s\,r$ and divided by $r$ to give entries, a layer's query-pairs can be calculated:
  \begin{equation}
    P^L = \underbrace{P(q, c; W)}_{\text{window}} \;+\; \frac{1}{r}\,\underbrace{P(q, c;\ s\,r)}_{\text{long range}},
  \end{equation}
  dropping the window term without a window, the kv cache term for a sliding-window layer, and the cap $s\,r$ without
  sparse selection. Every attention step is then modeled as a matmul, per layer:
  \begin{align}
    F_{QK}^L &= 2\,P^L H\, d_{qk}, \qquad F_{PV}^L = 2\,P^L H\, d_{v}, \\
    F_{idx}^L &= 2\,\tfrac{1}{r}\,P(q, c;\ N_{cand}\,r)\; H_{idx}\, d_{idx},
  \end{align}
  where $H_{idx}$ and $d_{idx}$ are the indexer's heads and key width, $N_{cand}$ is the number of entries it scores,
  and the indexer's pairs ignore the window and the top-$s$ cap. 

  \paragraph{KV cache} We model two aspects of the KV cache:
  \begin{enumerate}
      \item Capacity: how many bytes one request \emph{stores}, which sets the memory limit on concurrency
            $C_{\text{memory}}$
      \item Reads: how many bytes one decode step \emph{loads}, which is part of $D^L$
  \end{enumerate}
  These differ for sparse layers, so we compute them separately. For a context of $n_{ctx}$ tokens, let's define
  \begin{itemize}
      \item $n_{win} = \min(n_{ctx},\ W)$, the tokens held in the window
      \item $n_{kv} = \lceil n_{ctx} / r \rceil$, the entries of the kv cache
      \item $w_{kv}$, the values stored per entry: $H_{kv}(d_{qk} + d_v)$ for GQA ($H_{kv}\,d_{qk}$ with shared KV) and
            $d_c + d_r$ for MLA
      \item $d_{idx}$, the width of one index key, stored per kv cache entry for the sparse indexer
      \item $b_{kv}$, the bytes per element of the KV cache
  \end{itemize}

  \paragraph{Stored} A layer stores its window and its kv cache, plus an index key per kv cache entry if it is
  sparse:
  \begin{equation}
    S^L_{\text{stored}} = b_{kv} \cdot \big(n_{win} + n_{kv}\big) \cdot w_{kv} + b_{kv} \cdot n_{kv} \cdot d_{idx}.
  \end{equation}
  Some layers reuse an earlier layer's kv cache or index keys instead of keeping their own so they store only
  what they own.

  \paragraph{Read per decode step} A decode step reads the window and only the kv cache entries the query attends to,
  which is the top $s$ for a sparse layer:
  \begin{equation}
    S^L_{\text{read}} = b_{kv} \cdot \big(n_{win} + \min(n_{kv},\ s)\big) \cdot w_{kv}.
  \end{equation}
  A layer that runs the indexer also reads the index key of every entry it scores (all of them, or a shortlist of
  $N_{cand}$), adding $b_{kv} \cdot \min(n_{lv},\ N_{cand}) \cdot d_{idx}$. Sparse layers therefore read far less than
  they store.

  \subsection{Mixture of Experts}
  Similarly to attention, the MoE layer can be naively modeled fairly simply, but more complex modeling can account for
  effects which heavily impact the set of proposed sharding schemes. Some definitions that allow us to represent an MoE
  layer:
  \begin{itemize}
      \item $B$, the total number of tokens that engage in an MoE layer, before the top-k
      \item $E$, the number of total routed experts
      \item $k$, the number of experts each token routes to
      \item $EP_{\text{MoE}}$, the number of EP ranks, indexed by $j$
      \item $E_{\text{rank}} = E / EP_{\text{MoE}}$, the number of experts per rank
      \item $L_{\text{MoE}}$, the number of MoE layers in the model
      \item $\pi_e$, the probability a token picks expert $e$, such that $\sum_e \pi_e = k$
      \item $S_{\text{exp}}$, the total bytes of the routed experts per layer
      \item $d_{\text{lat}}$, an optional hidden dimension size used by latent MoE implementations
      \item $F_{\text{pair}}$, the FLOPs for one token through one routed expert
  \end{itemize}

  \paragraph{Modeling Expert Imbalance} Due to our interest in producing both a theoretical model and a model based on
  empirical measurements, we ultimately decide to use $\pi_e$ as input distribution for our modeling. That is, we enable
  both selecting a specific distribution to use as a theoretical model of the probability $\pi_e$ and measuring that
  distribution directly from a model run. For the theoretical model, we by default use a Zipf distribution $\pi_{(i)} = \min(1, c / i^a)$ for the $i$-th most popular expert, with $c$ set such that $\sum_i \pi_{(i)} = k$, $a = 0$ is uniform, and larger $a$ is more skewed. For the empirical model we measure $\pi_e$ per layer as the share of tokens routed to expert $e$.
  The expected number of a token's $k$ picks which are routed to rank $j$ is $$\mu_j = \sum_{e \in j} \pi_e$$ Then, the
  proportion of all $B \cdot k$ (token, expert) pairs which are assigned to rank $j$ is $$f_j = \mu_j / k$$ Treating tokens as routing independently, and a token's $k$ picks as independent of one another, the number of (token, expert) pairs which fall on rank $j$ is $$A_j \sim \text{Binomial}(B \cdot k,\ f_j)$$
  If $Z_j$ is the number of experts on rank $j$ which are routed to by at least one token, then $$\mathbb{E}[Z_j] = \sum_{e \in j} \big(1 - (1 - \pi_e)^B\big)$$

  We take the approximation that each expert is routed to independently with probability $R_e = 1 - (1 - \pi_e)^B$, so $Z_j$ is a sum of
Bernoulli($R_e$) over the experts on rank $j$. Because a layer waits for its slowest rank, we take the expected
maximum over ranks, treating ranks as independent:
\[
  \mathbb{E}\big[\max_j X_j\big] = \sum_{x \ge 0} \Big(1 - \prod_j F_j(x)\Big),
\]
where $F_j$ is the CDF of $A_j$ or $Z_j$. The routed experts' compute and memory time per layer under EP are then
\[
  t_{\text{comp}} = \frac{F_{\text{pair}} \cdot \mathbb{E}[\max_j A_j]}{\text{peak FLOP/s}}, \qquad
  t_{\text{mem}} = \frac{(S_{\text{exp}} / E) \cdot \mathbb{E}[\max_j Z_j]}{\text{HBM BW}},
\]
which enter the per-layer $\max(t_{\text{mem}}, t_{\text{comp}})$ alongside the MoE layer's shared and router weights.

\paragraph{Expert Imbalance: MoE TP}
Under MoE TP, there is no expert imbalance, because experts and tokens are sharded across the full TP domain.

  \subsection{Collectives and network}
  Every parallelism scheme except pure data parallelism requires chips to exchange activations within each layer.
  We model each exchange as one of four standard collectives over the $n$ chips that share the work, and compute its
  time $T^{L,i}_{\text{comm}}$ from the bytes $S$ each chip moves. Collectives are never overlapped with compute or
  memory reads: they are added to the layer's time.

  \paragraph{Which collectives each scheme needs} Per layer:
  \begin{center}
  \begin{tabular}{lll}
    \hline
    Parallelism & Collective & Over \\
    \hline
    Attention TP    & all-reduce of the output projection's partial sums & $n = TP_{\text{attn}}$ \\
    KV CP           & all-gather queries, reduce-scatter partial outputs   & $n = CP_{\text{kv}}$ \\
    Activation CP   & cheaper of: all-gather the KV, or all-gather queries + reduce-scatter outputs & $n = CP_{\text{act}}$ \\
    Dense TP        & all-gather the attention groups' activations, then all-reduce & $n = TP_{\text{dense}}$ \\
    Expert TP       & all-reduce of partial expert outputs & $n = N$ \\
    Expert parallel & all-to-all dispatch, then all-to-all combine & $n = N$ \\
    \hline
  \end{tabular}
  \end{center}

  \paragraph{Inside one scale-up domain} The chips connected by the scale-up fabric form a domain of $N_{dom}$ chips.
  A collective inside it is timed from the most specific data available for the chip:
  \begin{enumerate}
    \item \emph{Measured curves}: time versus message size from our collective micros interpolated
          log-log between measured sizes. Below the smallest measured size the time is constant, we determine this to be the call's fixed cost.
    \item \emph{Library fit}: a latency $\alpha$ and scale up fabric bandwidth $BW_{su}$ fitted per collective,
          \begin{equation}
            T^{L,i}_{\text{comm}} = \alpha + S \cdot \phi(n) / BW_{su},
          \end{equation}
          with nccl-tests' bus factors $\phi(n) = 2(n-1)/n$ for all-reduce and $(n-1)/n$ for the others.
    \item \emph{Link model} (spec sheet only): the bytes $S_{link}$ crossing the busiest link, of bandwidth
          $BW_{link}(n)$, plus a latency $\alpha_{hop}$ per hop over $h(n)$ hops,
          \begin{equation}
            T^{L,i}_{\text{comm}} = h(n)\cdot\alpha_{hop} \;+\; \frac{S_{link}}{BW_{link}(n)},
          \end{equation}
          where the topology sets $S_{link}$, $BW_{link}$ and the hop count $h$:
          \begin{itemize}
            \item \emph{Switch}: one hop at the chip's full bandwidth. A reduce-scatter sends the whole vector and the
                  switch reduces it; an all-gather receives $(n-1)/n$ of it.
            \item \emph{Full mesh}: a direct link to every other chip, each $1/(N_{dom}-1)$ of the chip's bandwidth, so
                  a collective over $n < N_{dom}$ chips uses only $n-1$ of them.
            \item \emph{Torus}: two links per axis, $h = \sum_{axis} \lfloor K_{axis}/2 \rfloor$ hops for the slice's
                  most cube-like shape with axis lengths $K_{axis}$, and the bytes are carried once per hop.
          \end{itemize}
  \end{enumerate}

  \paragraph{Across domains: the expert all-to-all} When expert parallelism spans more than one domain, each chip sends
  part of its tokens, $S_{NIC}$ bytes, over its NIC (bandwidth $BW_{NIC}$) and the rest over the scale-up links, taking
  $T_{\text{inside}}$. The two run at the same time, so the slower one sets the time:
  \begin{equation}
    T_{a2a} = \max\!\Big(T_{\text{inside}},\ \frac{S_{NIC}}{BW_{NIC}}\Big),
  \end{equation}
  or their sum on PCIe cards, where one link carries both. Two dispatch styles are modeled, one per phase:
  \begin{itemize}
    \item \emph{Decode} (low-latency kernels): each (token, expert) pair is sent directly to its chip.
    \item \emph{Prefill} (throughput kernels): a token crosses the NIC once per destination node and is then forwarded
          over that node's links; the combine sums expert outputs before sending them back.
  \end{itemize}

  \subsection{Customizability}
  Every input to the model is a separate, configurable parameter, from theoretical spec-sheet values to measured
  micro-benchmarks. Each can be swapped independently to trade simplicity for realism, and the best scheme and its cost
  are recomputed with every change.

  \paragraph{Theoretical and measured specs} Each chip is evaluated under four spec sources, from the vendor's spec sheet
  to measured values: 
  \begin{enumerate}
    \item \emph{theoretical}: the vendor spec sheet
    \item \emph{own micro}: our micro-benchmarks
    \item \emph{vendor tool}: vendor benchmarks (nvbandwidth, BabelStream, perftest, \dots)
    \item \emph{vendor library}: vendor libraries (cuBLASLt, hipBLASLt, nccl-tests, \dots)
  \end{enumerate}
  Comparing each tier allows us to identify peak achievable performance that is not hindered by software immaturity.
